\documentclass[12pt]{article}
\usepackage{graphicx}
\usepackage{amsmath}
\usepackage{amssymb}
\usepackage[square,numbers]{natbib}
\usepackage[subrefformat=parens]{subcaption}
\usepackage{booktabs}
\usepackage{url}
\usepackage[top=30truemm,bottom=30truemm,left=25truemm,right=25truemm]{geometry}
\begin{document}
\title{Supervised Contrastive Learning with Heterogeneous Similarity for Distribution Shifts}
\author{Takuro Kutsuna$\/^*$}
\date{\normalsize Toyota Central R\&D Labs., Inc. \\ $\/^*$kutsuna@mosk.tytlabs.co.jp}
\maketitle
\begin{abstract}
    Distribution shifts are problems where the distribution of data changes between training and testing, which can significantly degrade the performance of a model deployed in the real world.
    Recent studies suggest that one reason for the degradation is a type of overfitting, and that proper regularization can mitigate the degradation, especially when using highly representative models such as neural networks.
    In this paper, we propose a new regularization using the supervised contrastive learning to prevent such overfitting and to train models that do not degrade their performance under the distribution shifts.
    We extend the cosine similarity in contrastive loss to a more general similarity measure and propose to use different parameters in the measure when comparing a sample to a positive or negative example, which is analytically shown to act as a kind of margin in contrastive loss.
    Experiments on benchmark datasets that emulate distribution shifts, including subpopulation shift and domain generalization, demonstrate the advantage of the proposed method over existing regularization methods.    
\end{abstract}
\section{Introduction} \label{sec:intro}
Distribution shifts are problems where the distribution of data changes between training and testing, which can significantly degrade the performance of a prediction model deployed in the real world.
There may be a variety of possible distributional shifts, but the following two types have been identified~\cite{pmlr-v139-koh21a}: \emph{subpopulation shift} and \emph{domain generalization}.
Subpopulation shift refers to a situation where the existence of groups in the data is assumed and the group ratio changes between the training and test environments.
For example, in pedestrian detection, whereas the proportion of men wearing hats was low when the training data was collected, the trend changes after deployment and the proportion of men wearing hats increases significantly; note that  the groups here are defined by $\text{hat}\times \text{gender}$.
Domain generalization refers to a situation where the environment in which data is collected changes between training and testing. In the pedestrian detection example, it corresponds to using a detector in Asia that has been trained with datasets collected in the US and EU.

One specific situation where subpopulation shift becomes problematic is when there is a minority group in the training data, as reported in~\cite{sagawa2019distributionally}.
In particular, when using models with high representation capacity, such as deep neural networks (DNNs), a phenomenon has been reported where the generalization accuracy for some minority groups deteriorates while the overall accuracy increases as the training progresses.
If such a minority group were to change to a majority in the test environment, it would cause a significant drop in accuracy.
This phenomenon can be viewed as overfitting to the majority group in the training data.
In fact, \cite{sagawa2019distributionally} showed that applying a general regularization for DNNs during training prevents the minority group from deteriorating in accuracy.
Through our experiments with benchmark datasets that emulate domain generalization situations, we have observed phenomena similar to subpopulation shift.
That is, the accuracy for the domain(s) used in training improves as the training proceeds, while the accuracy for data in other domains not included in the training data gradually deteriorates.
If we think of this in the same way as the subpopulation shift case, this could be considered a kind of overfitting to the domain(s) used in training.
It is important to note that in this case the model is not overfitting in the traditional sense, since the test accuracy for the same domain as the training data has been improved by training.

In this paper, we propose a new regularization method to prevent such overfitting and to train DNNs that do not degrade their performance under the distribution shifts, including subpopulation shift and domain generalization.
The proposed method makes use of contrastive learning, which has achieved great success in the field of representation learning~\cite{pmlr-v119-chen20j,he2020momentum}.
Among them, we adopt a supervised approach~\cite{khosla2020supervised} that can deal directly with labeled data.
In contrastive learning, positive and negative examples are defined for each sample in the training data, and embeddings are learned to be close to positive examples and far from negative examples.
Cosine similarity is commonly used in contrastive loss as a similarity measure to compare the embeddings.
The proposed method extends this cosine similarity to a more general similarity measure.
An interesting point is that simply changing the similarity does not significantly change the learning results with contrastive loss.
In contrast, we propose to use different parameters in the measure when comparing a sample to a positive or negative example.
We show that such operation acts as a kind of margin in contrastive loss.
Experimental results suggest that the proposed method can learn models that are more robust to distribution shifts than those using basic cosine similarity.

The contributions of this paper are as follows:
\begin{itemize}
    \item We propose a regularization method to extend supervised contrastive loss to train DNNs that are robust to distribution shifts, showing that manipulating similarity in the loss is equivalent to establishing a margin.
    \item Using benchmark datasets, we show that overfitting can occur in the domain generalization setting as well as for the minority group in the subpopulation shift.
    \item We empirically show that the proposed method can mitigate performance degradation in both subpopulation shift and domain generalization settings. Experimental results also show that the proposed method is more effective than the conventional regularization for DNNs.
\end{itemize}
In Section~\ref{sec:related_work}, we review existing work on subpopulation shift and domain generalization. In Section~\ref{sec:preliminary}, we briefly introduce the loss used in contrastive learning as well as some notation.
We explain the proposed method in Section~\ref{sec:proposed} and present empirical results in Section~\ref{sec:experiments}.
Section~\ref{sec:conclusion} summarizes the work.

\section{Related Work} \label{sec:related_work}
\paragraph{Distributionally robust optimization and subpopulation shift.}
In contrast to Empirical Risk Minimization (ERM)~\cite{vapnik1999overview}, which assumes the same distribution for training and test data, Distributionally Robust Optimization (DRO) assumes a distribution set~$\mathcal{P}$ of inputs~$x$ in the test environment and optimizes a model to minimize the worst-case loss over~$\mathcal{P}$.
Although early studies of DRO considered an~$\epsilon$ ball from a training distribution as~$\mathcal{P}$, it was later found that DRO with such~$\mathcal{P}$ falls into the same model trained with the training distribution~\cite{pmlr-v80-hu18a}.
As a way to make DRO meaningful, Hu et al.~\cite{pmlr-v80-hu18a} suggested assuming a latent variable~$z$ for~$x$ and defining~$\mathcal{P}$ as dependent on~$z$.
By considering the group of~$x$ as~$z$, DRO can be a method to alleviate the problem of subpopulation shift.
There have been several researches that propose the definition of~$\mathcal{P}$ depending on~$z$ and an algorithm to minimize the worst-case loss over such~$\mathcal{P}$~\cite{sagawa2019distributionally,REx,oren2019distributionally,zhou2021examining}.
Another line of research tries to automatically find the latent group of~$x$ and to apply the DRO according to the group found~\cite{liu2021just,bao2021predict}. Recently, Zhang et al. proposed an approach to this line of work based on contrastive learning~\cite{pmlr-v162-zhang22z}.
An important finding reported in~\cite{sagawa2019distributionally} is that simply applying DRO may not improve generalization performance for worst groups when using complex models such as DNNs.
In order to prevent such problems, it is crucial to apply appropriate regularization when training DNNs, as discussed in~\cite{sagawa2019distributionally}.

\paragraph{Domain generalization and representation learning.}
Before we get into domain generalization, let's first briefly introduce domain adaptation.
Given datasets or models in source domains, \emph{domain adaptation} aims to obtain a model in a target domain for which labeled datasets are available.
\emph{Unsupervised domain adaptation} is a variant of domain adaptation in which only unlabeled data is provided for the target domain.
The goal of \emph{domain generalization} is also to get a good model for the target domain, but under the assumption that we have no data for the target domain.
In this sense, the target domain is also called as out-of-distribution~(OOD), while the source domains are called in-distribution~(ID).
As surveyed in~\cite{9847099}, various methods for domain generalization have been proposed, e.g. domain-adversarial learning~\cite{li2018deep}.

Representation learning is a method for learning embedding models that exhibit high performance when transferred to downstream tasks, and has made significant progress in recent years.
Representation learning and domain generalization are closely related.
If representation learning yields good embedding models for ID data, these models will also be effective for OOD data.
In fact, many methods originally developed for representation learning have been applied to domain generalization tasks, e.g. a method using jigsaw puzzles~\cite{carlucci2019domain}.

Contrastive learning is a type of representation learning that attempts to learn embeddings such that positive pairs stay close together while negative pairs are separated.
Hadsell et al.~\cite{1640964} proposed an early form of contrastive learning, which was later used for domain generalization in~\cite{motiian2017unified}.
Schroff et al.~\cite{7298682} proposed triplet loss, which was later used as part of the training loss for domain generalization in~\cite{dou2019domain}.
As discussed in~\cite{kim2021selfreg}, the performance of these contrastive approaches is sensitive to the selection of negative samples.
To alleviate this problem, contrastive learning methods that do not require negative samples have been proposed~\cite{grill2020bootstrap,chen2021exploring}, on the basis of which Kim et al.~\cite{kim2021selfreg} proposed a method for domain generalization.
Another line of work to address this problem is to use InfoNCE loss\footnote{This loss is also referred to as multi-class $N$-pair loss~\cite{sohn2016improved}.}~\cite{oord2018representation}, where multiple negative samples are used for each anchor sample.
The performance of representation learning with InfoNCE is shown to improve by increasing the number of negative samples used in the loss, which is mainly explored in self-supervised settings~\cite{chen2020simple,he2020momentum}.
Khosla et al.~\cite{khosla2020supervised} demonstrated the advantage of InfoNCE in ordinary supervised classification problems, which they called supervised contrastive learning (SupCon).
In this paper, we propose a method for domain generalization based on SupCon, where the similarity function is extended from the usual cosine similarity.
Recently, Sun et al.~\cite{sun2022out} proposed an OOD detection method based on SupCon, which is different from our scope of preventing the performance degradation in distribution shifts, including subpopulation shift and domain generalization.

\section{Preliminary} \label{sec:preliminary}
In this section, we briefly review contrastive learning with InfoNCE and provide the notations necessary to explain the proposed method.
\subsection{Contrastive Learning} \label{sec:contrastive_learning}
In contrastive learning, for an input~$x \in X$, given a set of positive examples~$\mathcal{S}_x^p$ that are similar in some sense to~$x$ and a set of negative examples~$\mathcal{S}_x^n$ that are not similar to~$x$, the embedding function~$f_\theta : X \to \mathbb{R}^m$ is learned by minimizing the following loss function.
\begin{align}
    \mathcal{L}_{\text{cl}}(\theta) & := - \frac{1}{\left|\mathcal{S}_x^p\right|} \sum_{x_p \in \mathcal{S}_x^p} \log \frac{e^{ \text{sim}\left(f_\theta(x), f_\theta(x_p)\right)/\tau }}{e^{ \text{sim}\left(f_\theta(x), f_\theta(x_p)\right)/\tau } + \displaystyle{\sum_{x_n \in \mathcal{S}_x^n}} e^{ \text{sim}\left(f_\theta(x), f_\theta(x_n)\right)/\tau }},  \label{eq:cl-loss}
\end{align}
where $\text{sim}(\cdot, \cdot)$ is a function to compute the similarity between the embedding vectors, usually the cosine similarity, and~$\tau$ is a temperature parameter.
In the above loss,~$f_\theta$ is learned so that the embedding of~$x$ is similar to the embedding of the positive example~$x_p$ compared to that of the negative example~$x_n$.

\subsection{Self-supervised Contrastive Learning}
Self-supervised contrastive learning~\cite{chen2020simple,he2020momentum} uses the input~$x$ to learn~$f_\theta$, under the assumption that the label~$y$ corresponding to~$x$ is not given.
As a positive example for~$x$,~$x$ itself is used after some data augmentation such as cropping or horizontal flipping has been applied.
Self-supervised contrastive learning usually generates only one positive example for each~$x$ ($\left|\mathcal{S}_x^p\right|=1$).
All images other than~$x$ (and any data augmentation to them) are considered negative examples.
It has been shown in the literature that increasing the size of~$\mathcal{S}_x^n$ is important to improve the performance of the contrastive loss~\cite{chen2020simple}.
However, if the size of~$\mathcal{S}_x^n$ is made too large, the computer will run out of memory, so learning is not possible in a typical computing environment.
Momentum Contrast (MoCo)~\cite{he2020momentum} is an approach to alleviate such a problem that effectively increases the size of~$\mathcal{S}_x^n$ using a queue of embeddings computed by the model that takes a moving average during learning.

\subsection{Supervised Contrastive Learning (SupCon)}
SupCon~\cite{khosla2020supervised} deals with a situation where a discrete label~$y$ is given for an input~$x$.
Examples other than~$x$ in the dataset with the same label~$y$ are considered as positive examples for~$x$, while examples with different labels are considered as negative examples for~$x$.
Therefore, embeddings with the same label are expected to gather in the embedding space.
To effectively increase the size of~$\mathcal{S}_x^n$, methods for self-supervised contrastive learning such as MoCo are also applicable to the supervised setting.
We applied MoCo in our experiments, in which both~$\mathcal{S}_x^p$ and~$\mathcal{S}_x^n$ are enlarged using a queue.

\begin{figure}[tbp]
    \begin{center}
        \includegraphics[width=0.9\linewidth,clip]{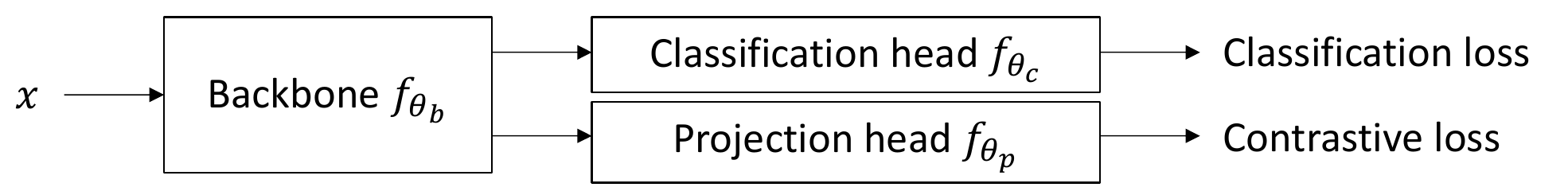}
        \caption{Model structure for supervised contrastive learning}
        \label{fig:supcon_model}
    \end{center}
\end{figure}

Since learning with the loss~(\ref{eq:cl-loss}) alone does not provide a model for predicting the label~$y$, a head model is prepared for classification as well as the projection head for contrastive loss.
Figure~\ref{fig:supcon_model} shows an overview of the model structure, in which~$f_{\theta_b}$,~$f_{\theta_c}$, and~$f_{\theta_p}$ denote the backbone model, the classification head, and the projection head, respectively.
The classification head~$f_{\theta_c}$ is a linear layer whose output is used in training with the classification loss, usually the cross entropy loss (CE).
The projection head~$f_{\theta_p}$ is introduced in self-supervised contrastive learning and is also employed in SupCon~\cite{khosla2020supervised}, for which we used a 3-layer multilayer perceptron following~\cite{chen2021empirical}.
The output of~$f_{\theta_p}$ is used as an embedding in the loss~(\ref{eq:cl-loss}), where~$f_\theta:=f_{\theta_p} \circ f_{\theta_b}$.

\begin{figure}[tbp]
    \centering
    \begin{minipage}[b]{0.3\linewidth}
        \centering
        \includegraphics[width=\linewidth,clip]{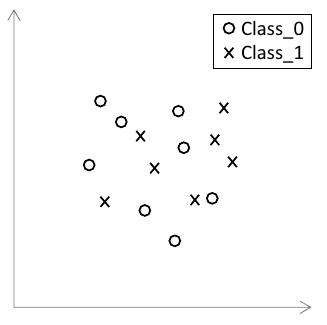}
        \subcaption{Initial embeddings.}
        \label{fig:ce_vs_supcon-init}
    \end{minipage} \hfill
    \begin{minipage}[b]{0.3\linewidth}
        \centering
        \includegraphics[width=\linewidth,clip]{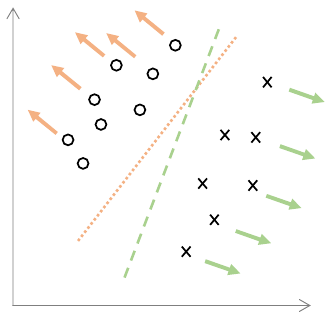}
        \subcaption{Training with CE.}
        \label{fig:ce_vs_supcon-ce}
    \end{minipage} \hfill
    \begin{minipage}[b]{0.3\linewidth}
        \centering
        \includegraphics[width=\linewidth,clip]{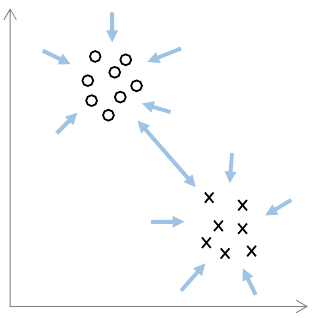}
        \subcaption{Training with SupCon.}
        \label{fig:ce_vs_supcon-supcon}
    \end{minipage}
    \caption{Sketch of the training with CE or SupCon: (a) Embedding space of the initial model. (b) Training with CE, where the dotted lines are the class-wise projections defined by the last linear layer. (c) Training with SupCon.  The arrows represent the forces induced by each loss.}
    \label{fig:ce_vs_supcon}
\end{figure}

Figure~\ref{fig:ce_vs_supcon} illustrates samples in the embedding space during training, showing the difference between CE and SupCon.
In the training with CE, the embedding function is learned so that each example is as far away from the current classification plane as possible, as shown in Figure~\ref{fig:ce_vs_supcon-ce}.
In contrast, training with SupCon forces samples of the same class to aggregate while keeping samples of different classes apart, as shown in Figure~\ref{fig:ce_vs_supcon-supcon}.
Therefore, one of the major differences between CE and SupCon is whether or not samples of the same class are intentionally being gathered.
In other words, SupCon tries to learn embeddings that are invariant for the same class, while preserving the features necessary to distinguish classes, which is expected to result in an embedding space that better represents the characteristics unique to each class, leading to a better generalization ability.
How well the embedding is aggregated for each class is relative to the characteristics of the class, as our experiments suggest.

\section{Proposed Method} \label{sec:proposed}
We propose a novel regularization method for SupCon by extending the similarity function used in the loss, for which we use the t-vMF similarity~\cite{tvMF}, which generalizes the cosine similarity.

\subsection{t-vMF similarity}
\begin{figure}[tbp]
    \centering
    \begin{minipage}[b]{0.49\linewidth}
        \centering
        \includegraphics[width=\linewidth,clip]{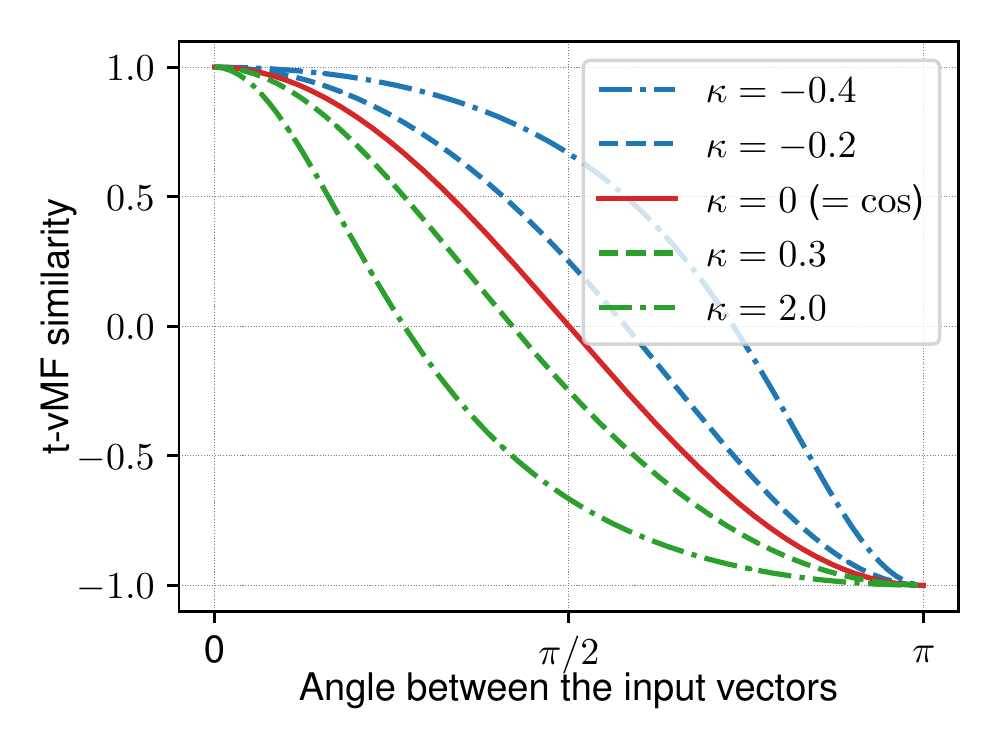}
        \caption{Plots of t-vMF similarity.}
        \label{fig:tvMF}
    \end{minipage}
    \begin{minipage}[b]{0.49\linewidth}
        \centering
        \includegraphics[width=\linewidth,clip]{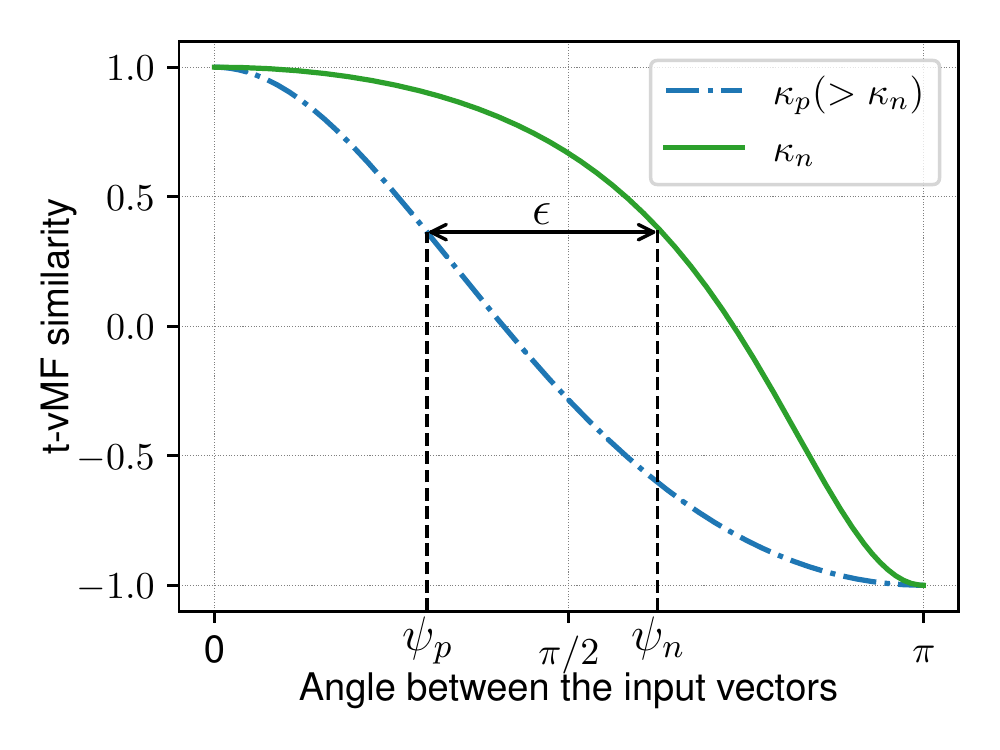}
        \caption{The margin~$\epsilon$ for $\kappa_p > \kappa_n$.}
        \label{fig:PNtvMF}
    \end{minipage}
\end{figure}

The t-vMF similarity is obtained by extending the von Mises-Fisher (vMF) distribution, which is a normal distribution on the hypersphere, to be heavy-tailed, named for its analogy to the student-t extension of the normal distribution.
The t-vMF similarity for two $m$-dimensional vectors~$a, b$ is given by
\begin{align}
    \phi_\kappa (a, b) := \frac{1+ \cos(a,b)}{1+\kappa\left(1-\cos(a,b)\right)} - 1, \label{eq:tvMF}
\end{align}
where~$\kappa \in (-1/2, \infty)$ is the hyperparameter of the t-vMF similarity and $\cos$ is the cosine similarity.
From the definition we see that the t-vMF similarity is equal to the cosine similarity when~$\kappa=0$.
Figure~\ref{fig:tvMF} shows an example of t-vMF similarity plots for $\kappa \in \left\{-0.4, -0.2, 0, 0.3, 2.0\right\}$, where the $x$-axis is the angle between the input vectors.
We see that the similarity is~$1$ when the angle is~$0$ and is~$-1$ when the angle is~$\pi$, regardless of the value of~$\kappa$.
It can also be seen that for~$\kappa=0$, which corresponds to cosine, the similarity is~$0$ at the angle~$\pi/2$, while for~$\kappa < 0$ the similarity is positive at the angle~$\pi/2$ and negative for~$\kappa > 0$.
It holds that $\phi_{\kappa_1} (a, b) \geq \phi_{\kappa_2} (a, b)$ for $\kappa_1 < \kappa_2$, where the equality holds only if the angle is~$0$ or~$\pi$.

\subsection{Proposed Loss}
We propose the following loss using the t-vMF similarity as the similarity in the contrastive loss~(\ref{eq:cl-loss}):
\begin{align}
    \tilde{\mathcal{L}}_{\text{cl}}(\theta) & := - \frac{1}{\left|\mathcal{S}_x^p\right|} \sum_{x_p \in \mathcal{S}_x^p} \log \frac{e^{ \phi_{\kappa_p}\left(f_\theta(x), f_\theta(x_p)\right)/\tau }}{e^{ \phi_{\kappa_p}\left(f_\theta(x), f_\theta(x_p)\right)/\tau } + \displaystyle{\sum_{x_n \in \mathcal{S}_x^n}} e^{ \phi_{\kappa_n}\left(f_\theta(x), f_\theta(x_n)\right)/\tau }}  \label{eq:cl-tvMF}
\end{align}
The important point is to use different~$\kappa$ for the positive and negative examples, $\kappa_p$ and~$\kappa_n$ respectively.
Interestingly, as our experiments suggest, there was no significant difference in test accuracy between using cosine similarity and t-vMF similarity as long as $\kappa_p=\kappa_n$.
In contrast, when $\kappa_p > \kappa_n$, there was a clear difference in the results, leading to improved accuracy under distribution shifts in our experiments.
The reason for this difference can be examined as follows.
We first transform the loss~(\ref{eq:cl-tvMF}) as
\begin{align}
    \tilde{\mathcal{L}}_{\text{cl}}(\theta) & = - \frac{1}{\left|\mathcal{S}_x^p\right|} \sum_{x_p \in \mathcal{S}_x^p} \log \frac{1}{1 + \displaystyle{\sum_{x_n \in \mathcal{S}_x^n}} e^{\left(\phi_{\kappa_n}\left(f_\theta(x), f_\theta(x_n)\right) - \phi_{\kappa_p}\left(f_\theta(x), f_\theta(x_p)\right)\right) / \tau }}.  \label{eq:cl-tvMF-mod}
\end{align}
If we define $\delta := \phi_{\kappa_n}\left(f_\theta(x), f_\theta(x_n)\right)- \phi_{\kappa_p}\left(f_\theta(x), f_\theta(x_p)\right)$,
it is necessary to make~$\delta$ lower to reduce~$\tilde{\mathcal{L}}_{\text{cl}}$.
Now we denote the angles that $x$ makes with~$x_p$ and~$x_n$ as~$\psi_p$ and~$\psi_n$, respectively, then the following holds for $\kappa_p=\kappa_n$:
\begin{align*}
    \psi_n - \psi_p > 0 & \quad \to \quad \delta<0, \\
    \psi_n - \psi_p = 0 & \quad \to \quad \delta=0, \\
    \psi_n - \psi_p < 0 & \quad \to \quad \delta>0.
\end{align*}
That is, $\delta=0$ if the angles with the positive and negative examples are the same, and $\delta < 0$ if the angle with the positive example is less than the angle with the negative example, indicating that $x$ is more similar to the positive example than the negative example.
In contrast, the following holds for $\kappa_p>\kappa_n$:
\begin{align*}
    \psi_n - \psi_p > \epsilon & \quad \to \quad \delta<0, \\
    \psi_n - \psi_p = \epsilon & \quad \to \quad \delta=0, \\
    \psi_n - \psi_p < \epsilon & \quad \to \quad \delta>0,
\end{align*}
where $\epsilon \ (> 0)$ is a non-negative margin whose value depends on~$\kappa_p$,~$\kappa_n$,~$\psi_p$, and~$\psi_n$.
Figure~\ref{fig:PNtvMF} illustrates an example of such~$\epsilon$.
Therefore, if $\kappa_p>\kappa_n$, $\psi_p$ must be less than~$\psi_n$ by some positive margin~$\epsilon$ to make $\delta < 0$.
Such a margin will act as a sort of regularization in the contrastive loss, leading to better accuracy under distribution shift.

A similar margin has been proposed in the literature for training with CE.
ArcFace~\cite{arcface} is a method for face recognition tasks, where a fixed margin is introduced in the loss.
LDAM~\cite{ldam} proposed a loss for dealing with the long-tail problem, in which class-dependent fixed margins are added.
The differences between the proposed method and these methods are as follows:
First, the proposed method is applied to the contrastive loss, whereas the existing methods are all applied to the CE. Second, the proposed method uses a variable margin based on t-vMF similarity, whereas the existing methods use fixed margins.

\paragraph{Controlling~$\kappa_p$ and~$\kappa_n$ with a single parameter.}
It is practically inconvenient to control~$\kappa_p$ and~$\kappa_n$ separately under~$\kappa_p>\kappa_n$.
To alleviate this problem, we propose to control~$\kappa_p$ and~$\kappa_n$ with a single parameter~$\alpha \in [0, 0.5)$, using the following equations:
\begin{align*}
    \kappa_p = \frac{\alpha}{2 \alpha + 1}, \quad \kappa_n = -\alpha
\end{align*}
These equations were derived from the condition\footnote{$1/(1+\kappa_p)-1 = -(1/(1+\kappa_n) - 1)$} that the similarity with~$\kappa_p$ and~$\kappa_n$ becomes symmetric at the angle~$\pi/2$.
For example, $\alpha=0.4$ results in~$(\kappa_p, \kappa_n)=(2.0, -0.4)$, which are shown in Figure~\ref{fig:tvMF}.
Note that as we set~$\alpha$ larger, $\kappa_p - \kappa_n$ becomes larger, which corresponds to a stronger regularization in our method.

\section{Experimental Results} \label{sec:experiments}
We demonstrate the usefulness of the proposed method using benchmark datasets that emulate subpopulation shift and domain generalization.

\subsection{Datasets}
To evaluate a subpopulation shift situation, we use the CelebA dataset~\cite{celaba}, which consists of face images with multiple annotations.
Following~\cite{sagawa2019distributionally}, we consider the problem of predicting whether the hair color of a person in an image is blond or not\footnote{
    We ran another experiment with the CelebA dataset, where the goal is to predict whether a person in an image is smiling or not. We used ``mouth\_slightly\_open'' as a confounding variable. The results are reported in Appendix~\ref{appendix:celeba}.}.
Each training image belongs to one of four groups defined by hair color and gender, whose contingency table is given in Table~\ref{tab:celeba}, from which we can see that blond hair males are a minority.
The test split of the CelebA dataset contains about 20000 test images, and the percentage of data contained in them is also about the same as in Table~\ref{tab:celeba}.

We use the Camelyon17-WILDS dataset~\cite{pmlr-v139-koh21a},
which we will refer to as Camelyon17 for simplicity, to evaluate a situation of domain generalization.
Camelyon17 consists of histopathological images 
collected from five hospitals.
Our goal here is to predict the label~$y$ that indicates whether the image contains tumor tissue or not.
The Camelyon17 dataset is divided into four splits\footnote{These splits are named train, id\_val, val, and test in the original data.}, which we call $\mathcal{D}_\text{train}$, $\mathcal{D}_\text{test}^\text{ID}$, $\mathcal{D}_\text{test}^\text{OOD1}$, and $\mathcal{D}_\text{test}^\text{OOD2}$.
Table~\ref{tab:camelyon17} summarizes from which hospital each of the splits was obtained, where hospitals are referred to as centers.
It is well known that microscopic images have different trends depending on the hospital due to microscope types and staining methods~\cite{KOMURA201834}.
Note that the label~$y$ is adjusted to be uniformly included in each split.

\begin{table}[tbp]
    \centering
    \caption{Contingency tables of the datasets used in the experiment.}
    \label{tab:contingency}
    \tabcolsep = 4pt
    \begin{minipage}[b]{0.44\hsize}
        \centering
        \subcaption{CelebA (training split)}
        \label{tab:celeba}
        \scalebox{0.9}[0.9]{
            \begin{tabular}{lrr}
                \toprule
                                 & Male  & Female \\
                \midrule
                Not\_Blond\_Hair & 66874 & 71629  \\
                Blond\_Hair      & 1387  & 22880  \\
                \bottomrule
            \end{tabular}
        }
    \end{minipage} \hfill
    \begin{minipage}[b]{0.55\hsize}
        \centering
        \subcaption{Camelyon17}
        \label{tab:camelyon17}
        \scalebox{0.9}[0.9]{
            \begin{tabular}{crrrr}
                \toprule
                Center & $\mathcal{D}_\text{train}$ & $\mathcal{D}_\text{test}^\text{ID}$ & $\mathcal{D}_\text{test}^\text{OOD1}$ & $\mathcal{D}_\text{test}^\text{OOD2}$ \\
                \midrule
                0      & 53425                      & 6011                                & 0                                     & 0                                     \\
                1      & 0                          & 0                                   & 34904                                 & 0                                     \\
                2      & 0                          & 0                                   & 0                                     & 85054                                 \\
                3      & 116959                     & 12879                               & 0                                     & 0                                     \\
                4      & 132052                     & 14670                               & 0                                     & 0                                     \\
                \bottomrule
            \end{tabular}
        }
    \end{minipage}
\end{table}

\subsection{Model and training settings}
We used ResNet18~\cite{resnet} as the backbone model in Figure~\ref{fig:supcon_model}.
When applying SupCon, we adopted MoCo~\cite{he2020momentum}, where the queue size was set to 65536 according to~\cite{he2020momentum}.
Unless otherwise stated, the temperature~$\tau$ in the SupCon loss is set to 1 and the loss coefficient for both the classification loss and the SupCon loss is 1.
As the classification loss for CelebA, we applied two types of group DRO losses, the Robust-DRO loss~\cite{sagawa2019distributionally} and the CVaR-DRO loss~\cite{oren2019distributionally}, using four groups defined by hair color and gender in Table~\ref{tab:celeba}.
We used the implementation\footnote{\url{https://github.com/kohpangwei/group_DRO} (07/03/2023)} by the authors of Robust-DRO and the implementation\footnote{\url{https://github.com/violet-zct/group-conditional-DRO} (07/03/2023)} by the authors of the paper~\cite{zhou2021examining} related to CVaR-DRO.
In training with group DRO losses, we resampled samples to balance groups in a minibatch according to~\cite{sagawa2019distributionally}.
For Camelyon17, we used CE as the classification loss.

The Adam optimizer~\cite{adam} was used for training, and 50 epochs were trained for CelebA and 25 epochs for Camelyon17, with an initial learning rate of 0.001.
Unless otherwise stated, the learning rate was decayed during training using the cosine schedule without restarts~\cite{loshchilovsgdr} and the weight decay was set to 0 .
Random cropping and random horizontal flipping were used as data augmentation, except for the experiments with RandAugment~\cite{randaugment}.
No pretrained models were used in the experiments.
Training was performed 5~times in each setting with different random seeds, and we evaluated the results with the mean and standard deviation of them.
All experiments were run using MMClassification~\cite{2020mmclassification}.

\subsection{Results for subpopulation shift} \label{sec:res_celeba}
\begin{figure}[tb]
    \centering
    \includegraphics[width=1.0\linewidth,clip]{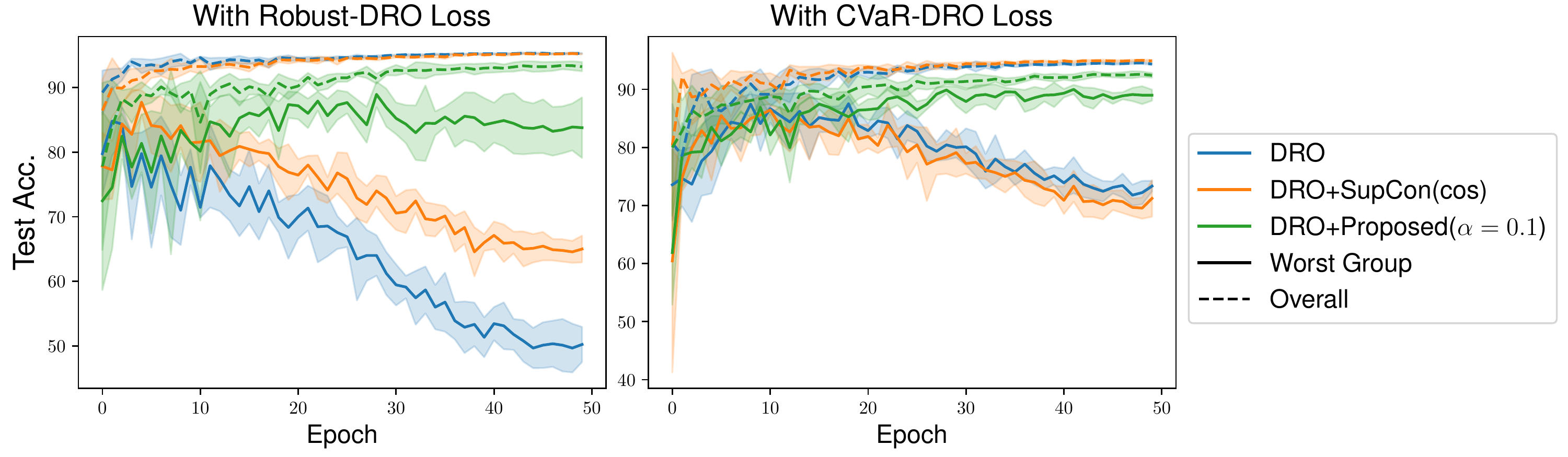}
    \caption{Test accuracy of the worst group (solid) and overall (dashed) for CelabA.}
    \label{fig:celeba_r-dro_cvar-dro}
\end{figure}

We first compare three models using the CelebA dataset, a model trained with the group DRO loss, a model trained with the group DRO loss and the SupCon loss with cosine similarity, and a model trained with the group DRO loss and the proposed loss with~$\alpha=0.1$ explained in Section~\ref{sec:proposed}.
The test accuracy curves during training are shown in Figure~\ref{fig:celeba_r-dro_cvar-dro}, where the solid and dashed lines represent the worst group and overall accuracy, respectively.
Note that, as discussed in Section~\ref{sec:intro}, it is important to improve the worst group accuracy under the subpopulation shift assumption.
From Figure~\ref{fig:celeba_r-dro_cvar-dro} we can see that the worst group accuracy degrades as training progresses, even though group DRO losses are applied; this behavior is consistent with that reported in~\cite{sagawa2019distributionally}.
In contrast, we can prevent the worst group loss from degrading by applying the proposed loss, regardless of the type of DRO loss used.
SupCon loss with the ordinary cosine similarity can slightly mitigate the degradation when used with Robust-DRO loss, but has little effect when used with CVaR-DRO.

\begin{figure}[tb]
    \centering
    \includegraphics[width=0.65\linewidth,clip]{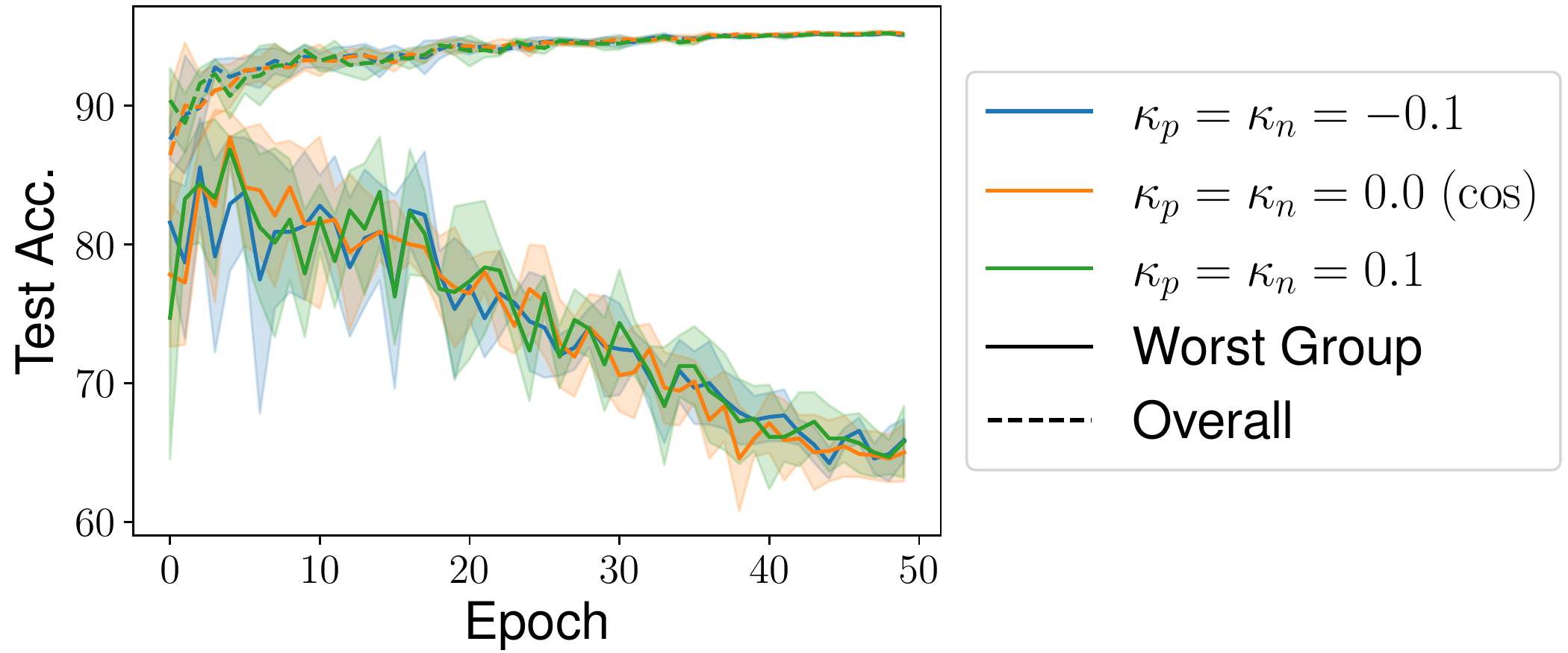}
    \caption{Test accuracy for CelebA when $\kappa_p = \kappa_n$, trained with Robust-DRO loss.}
    \label{fig:celeba_kp_eq_kn}
\end{figure}

To investigate the importance of using~$\kappa_p$ and~$\kappa_n$ such that~$\kappa_p \neq \kappa_n$ in the proposed loss, we trained models with Robust-DRO loss and the loss~(\ref{eq:cl-tvMF}) with $\kappa_p = \kappa_n$.
Figure~\ref{fig:celeba_kp_eq_kn} shows the results, indicating that simply using t-vMF similarity as the similarity measure for SupCon loss has little effect compared to using the cosine similarity.
It is essential to use different~$\kappa$ for the positive and negative samples, which leads to a kind of margin effect as discussed in Section~\ref{sec:proposed}.

We compared the proposed method with standard regularization methods for DNNs: weight decay~(WD), stochastic depth~(SD)~\cite{stoc_depth}, and RandAugment~\cite{randaugment}.
Although SD is only applicable to DNN models with specific structures, such as ResNet, it has been shown in the literature to be an effective regularization~\cite{xie2020self}.
RandAugment is a data augmentation method rather than a regularization method, but it is expected to prevent overfitting by strong data augmentation.
We used the implementation in~\cite{2020mmclassification} for these methods.
Table~\ref{tab:res_blond_gender_worst_group} summarizes the results, evaluating the test accuracy of the worst group at the end of training\footnote{The overall accuracy is summarized in Appendix~\ref{appendix:celeba}.}.
We compared two types of DRO losses mentioned above, as well as the learning rate scheduling including the fixed rate and the cosine decay.
From Table~\ref{tab:res_blond_gender_worst_group}, we can see that the accuracy of the worst group is improved by applying the regularization method compared to the model trained using only the DRO loss.
The proposed method achieves comparable or better results than existing regularization methods, independent of learning rate and loss type.
\begin{table}[tb]
    \center
    \tabcolsep = 4pt
    \caption{Test accuracy of the worst group for CelebA.}
    \label{tab:res_blond_gender_worst_group}
    \scalebox{0.9}[0.9]{
        \begin{tabular}{lllllr}
            \toprule
                             & \multicolumn{2}{l}{Fixed learning rate} & \multicolumn{2}{l}{Learning rate with decay} &                                                 \\
            \cmidrule(lr){2-3}\cmidrule(lr){4-5}
            {}               & Robust-DRO                              & CVaR-DRO                                     & Robust-DRO        & CVaR-DRO          & Average \\
            \midrule
            DRO-Loss         & 54.44 ($\pm$3.82)                       & 67.56 ($\pm$8.04)                            & 50.22 ($\pm$2.71) & 73.33 ($\pm$0.93) & 61.39   \\
            \addlinespace
            + WD(1e-4)       & 78.89 ($\pm$6.97)                       & 79.62 ($\pm$6.85)                            & 65.44 ($\pm$2.49) & 85.56 ($\pm$1.05) & 77.38   \\
            + WD(1e-3)       & 79.71 ($\pm$6.73)                       & 75.50 ($\pm$14.56)                           & 81.33 ($\pm$3.12) & 88.33 ($\pm$0.86) & 81.22   \\
            + WD(1e-2)       & 82.26 ($\pm$4.97)                       & 61.28 ($\pm$13.84)                           & 87.33 ($\pm$2.47) & 88.97 ($\pm$1.44) & 79.96   \\
            \addlinespace
            + SD(0.1)        & 60.44 ($\pm$2.29)                       & 76.89 ($\pm$6.01)                            & 58.56 ($\pm$1.88) & 77.78 ($\pm$2.25) & 68.42   \\
            + SD(0.3)        & 73.22 ($\pm$4.06)                       & 84.00 ($\pm$3.40)                            & 74.44 ($\pm$1.22) & 85.22 ($\pm$1.25) & 79.22   \\
            + SD(0.5)        & 82.78 ($\pm$5.01)                       & 85.78 ($\pm$5.15)                            & 82.67 ($\pm$0.42) & 88.56 ($\pm$0.75) & 84.95   \\
            \addlinespace
            + RandAug        & 78.33 ($\pm$4.47)                       & 87.27 ($\pm$1.57)                            & 76.56 ($\pm$0.96) & 90.11 ($\pm$1.12) & 83.07   \\
            \addlinespace
            + SupCon(cos)    & 74.89 ($\pm$5.36)                       & 73.44 ($\pm$4.19)                            & 65.00 ($\pm$2.08) & 71.22 ($\pm$3.17) & 71.14   \\
            \addlinespace
            + Proposed(0.05) & 83.44 ($\pm$2.03)                       & 84.78 ($\pm$3.31)                            & 79.56 ($\pm$2.75) & 80.56 ($\pm$0.99) & 82.09   \\
            + Proposed(0.1)  & 86.56 ($\pm$0.96)                       & 82.90 ($\pm$4.14)                            & 83.78 ($\pm$4.69) & 89.00 ($\pm$0.96) & 85.56   \\
            + Proposed(0.15) & 84.67 ($\pm$2.91)                       & 85.35 ($\pm$4.81)                            & 88.44 ($\pm$0.96) & 89.24 ($\pm$1.47) & 86.93   \\
            + Proposed(0.2)  & 85.46 ($\pm$2.24)                       & 80.74 ($\pm$1.27)                            & 85.78 ($\pm$1.47) & 88.66 ($\pm$1.94) & 85.16   \\
            \bottomrule
        \end{tabular}
    }
\end{table}

We also compared the proposed method with the SupCon loss with cosine similarity by varying several parameters.
First, we increased the loss weight for SupCon when combined with a DRO loss to enhance the effect of SupCon in training.
Next, we changed the temperature~$\tau$.
We also implemented the SupCon with a fixed margin similar to~\cite{arcface}, which replaces~$\cos(\theta_p)$ with~$\cos(\theta_p + m)$, where~$\theta_p$ is the angle between~$f_\theta(x)$ and~$f_\theta(x_p)$ in~(\ref{eq:cl-loss}) and $m$ is the fixed margin. Note that~$m=0.5$ is used in~\cite{arcface}.
The results are summarized in Table~\ref{tab:res_blond_gender_vs_other_supcon}, where the training were done with Robust DRO loss.
We can see that the proposed method performs better than the other settings for~$\alpha \geq 0.1$.
\begin{table}[tbp]
    \center
    \tabcolsep = 10pt
    \caption{Comparison with other SupCon settings}
    \label{tab:res_blond_gender_vs_other_supcon}
    \scalebox{0.9}[0.9]{
        \begin{tabular}{ll}
            \toprule
            {}                   & WorstGroup        \\
            \midrule                                   
            SupCon(cos)          & 65.00 ($\pm$2.08) \\
            \addlinespace                              
            + loss\_weight(5)    & 73.89 ($\pm$1.27) \\
            + loss\_weight(10)   & 76.11 ($\pm$2.25) \\
            \addlinespace                              
            + temperature(0.1)   & 43.56 ($\pm$0.97) \\
            + temperature(10)    & 76.78 ($\pm$4.24) \\
            \addlinespace                              
            + fixed\_margin(0.1) & 81.00 ($\pm$1.08) \\
            + fixed\_margin(0.3) & 81.78 ($\pm$5.34) \\
            + fixed\_margin(0.5) & 83.11 ($\pm$4.41) \\
            \addlinespace                              
            Proposed(0.05)       & 79.56 ($\pm$2.75) \\
            Proposed(0.1)        & 83.78 ($\pm$4.69) \\
            Proposed(0.15)       & 88.44 ($\pm$0.96) \\
            Proposed(0.2)        & 85.78 ($\pm$1.47) \\
            \bottomrule
        \end{tabular}
    }
\end{table}

\subsection{Results for domain generalization}
\begin{figure}[tb]
    \centering
    \includegraphics[width=1.0\linewidth,clip]{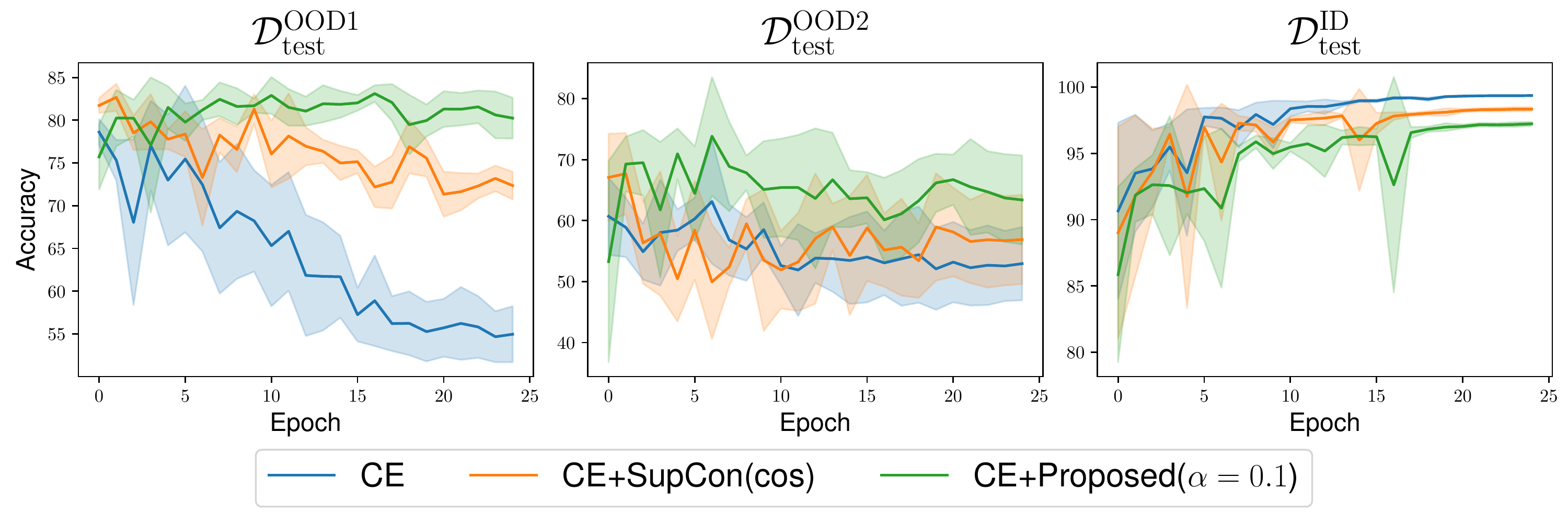}
    \caption{Accuracy on the test splits of the Camelyon17 dataset.}
    \label{fig:camelyon17_vs_cos}
\end{figure}

We trained three models using~$\mathcal{D}_\text{train}$ from the Camelyon17 dataset: a model trained with CE, CE and SupCon with cosine similarity, and CE and the proposed loss with~$\alpha=0.1$.
The trained models are evaluated using each of~$\mathcal{D}_\text{test}^\text{OOD1}$, $\mathcal{D}_\text{test}^\text{OOD2}$, and~$\mathcal{D}_\text{test}^\text{ID}$.
Figure~\ref{fig:camelyon17_vs_cos} shows the test accuracy curves during training.
Interestingly, the model trained with CE achieves high accuracy on~$\mathcal{D}_\text{test}^\text{ID}$, while the accuracy on~$\mathcal{D}_\text{test}^\text{OOD1}$ and~$\mathcal{D}_\text{test}^\text{OOD2}$ degrades as the training progresses.
This behavior is similar to the worst group accuracy when training with CelebA.
Note that such performance degradation is different from what is commonly referred to as overfitting, because the model achieves high accuracy on~$\mathcal{D}_\text{test}^\text{ID}$.
From Figure~\ref{fig:camelyon17_vs_cos} we can see that the performance degradation on the OOD datasets can be prevented by the proposed method, while the accuracy on~$\mathcal{D}_\text{test}^\text{ID}$ got slightly worse when trained with the proposed loss.

We compared the proposed method with other regularization methods using the Camelyon17 dataset.
Table~\ref{tab:res_camelyon17} summarizes the results, where the accuracy on each test split is evaluated at the end of training.
We can see that the accuracy of the OOD splits is improved by applying the regularization, which is analogous to the improvement of the worst group accuracy for the CelebA dataset.
The proposed method achieves comparable or better performance on OOD splits than existing regularization methods, especially on~$\mathcal{D}_\text{test}^\text{OOD1}$, whose performance deteriorates significantly when trained without regularization.
\begin{table}[tb]
    \center
    \tabcolsep = 10pt
    \caption{Comparison of accuracy on the test splits of the Camelyon17 dataset.}
    \label{tab:res_camelyon17}
    \scalebox{0.9}[0.9]{
        \begin{tabular}{llll}
            \toprule
            {}               & $\mathcal{D}_\text{test}^\text{OOD1}$ & $\mathcal{D}_\text{test}^\text{OOD2}$ & $\mathcal{D}_\text{test}^\text{ID}$ \\
            \midrule
            CE               & 54.97 ($\pm$3.28)                     & 52.94 ($\pm$6.01)                     & 99.36 ($\pm$0.03)                   \\
            \addlinespace
            + WD(1e-4)       & 60.50 ($\pm$1.61)                     & 68.13 ($\pm$8.56)                     & 99.00 ($\pm$0.06)                   \\
            + WD(1e-3)       & 62.17 ($\pm$7.29)                     & 59.23 ($\pm$5.42)                     & 98.57 ($\pm$0.06)                   \\
            + WD(1e-2)       & 79.97 ($\pm$3.06)                     & 59.91 ($\pm$9.10)                     & 97.87 ($\pm$0.10)                   \\
            \addlinespace
            + SD(0.1)        & 55.22 ($\pm$2.63)                     & 58.91 ($\pm$2.88)                     & 99.31 ($\pm$0.01)                   \\
            + SD(0.3)        & 58.32 ($\pm$2.54)                     & 63.13 ($\pm$3.96)                     & 99.20 ($\pm$0.04)                   \\
            + SD(0.5)        & 67.15 ($\pm$4.89)                     & 64.17 ($\pm$2.59)                     & 98.94 ($\pm$0.06)                   \\
            \addlinespace
            + RandAug        & 72.69 ($\pm$4.27)                     & 66.85 ($\pm$2.58)                     & 98.92 ($\pm$0.03)                   \\
            \addlinespace
            + SupCon(cos)    & 72.35 ($\pm$1.62)                     & 56.89 ($\pm$7.32)                     & 98.33 ($\pm$0.15)                   \\
            \addlinespace
            + Proposed(0.05) & 74.63 ($\pm$3.49)                     & 55.21 ($\pm$6.84)                     & 97.40 ($\pm$0.21)                   \\
            + Proposed(0.1)  & 80.24 ($\pm$2.39)                     & 63.39 ($\pm$7.30)                     & 97.22 ($\pm$0.16)                   \\
            + Proposed(0.15) & 80.18 ($\pm$1.87)                     & 59.45 ($\pm$10.02)                    & 97.01 ($\pm$0.39)                   \\
            + Proposed(0.2)  & 79.66 ($\pm$1.27)                     & 53.70 ($\pm$11.89)                    & 97.21 ($\pm$0.31)                   \\
            \bottomrule
        \end{tabular}
    }
\end{table}

\begin{figure}[tb]
    \centering
    \includegraphics[width=1.0\linewidth,clip]{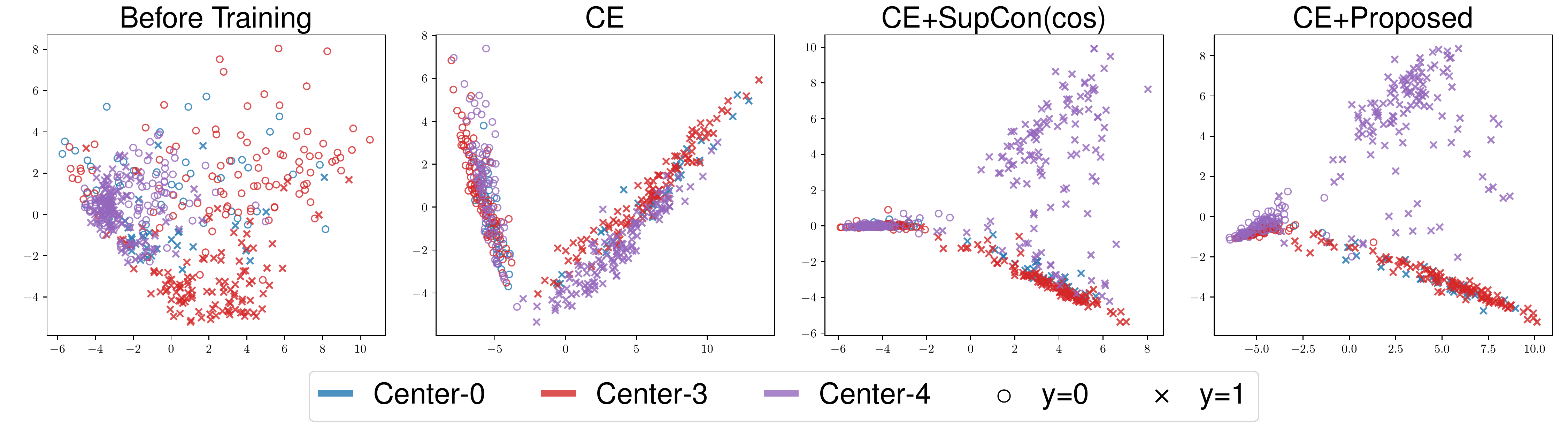}
    \caption{PCA visualization of the embeddings for samples in~$\mathcal{D}_{\text{train}}$.}
    \label{fig:camelyon17_pca_train}
\end{figure}

To investigate the effect of training with SupCon combined with CE, we visualized the embeddings obtained by the backbone model using Principal Component Analysis (PCA).
Figure~\ref{fig:camelyon17_pca_train} shows the embeddings for samples in~$\mathcal{D}_{\text{train}}$, where the point colors correspond to the hospital where the sample was obtained, and the ``o'' and ``x'' marks mean~$y=0$ (no tumor) and~$y=1$ (tumor), respectively\footnote{Visualizations of embeddings from test splits are shown in Appendix~\ref{appendix:camelyon17}.}.
We can see that the embeddings by the model trained with CE only are separated in the embedding space to discriminate the label.
In contrast, the embeddings obtained by CE and SupCon, or CE and proposed, are more aggregated between the same class, except for the~$y=1$ samples from Center-4, which may be due to the characteristics of the tumor samples from Center-4 are different from those from other centers.
The results in Figure~\ref{fig:camelyon17_pca_train} are consistent with the discussion in Figure~\ref{fig:ce_vs_supcon}.

\paragraph{Limitations of the proposed method.}
We also conducted experiments using the iWildCam dataset~\cite{pmlr-v139-koh21a} as another dataset to evaluate domain generalization.
Compared to the results for Camelyon17, we find little difference in OOD performance between models trained with CE, CE and SupCon, or CE and proposed.
Unlike Camelyon17, we did not observe any performance degradation on OOD splits from iWildCam during training.
The reason for this is not clear, but the training split of iWildCam includes samples from many locations corresponding to hospitals in Camelyon17, which may have led to the prevention of overfitting.
These results suggest that the proposed method has an effect of mitigating performance degradation on OOD splits, but has little effect when no performance degradation occurs, neither bad side effects.
See Appendix~\ref{appendix:iwildcam} for details.

\section{Conclusion} \label{sec:conclusion}
We have proposed a novel method to regularize DNN training to improve the generalization ability under distribution shifts. The proposed method extends the supervised contrastive loss using the t-vMF similarity, which generalizes the cosine similarity. We analytically investigated that we can establish a kind of margin in the contrastive loss by using different similarity parameters for positive and negative examples.
Experimental results suggest that it is essential to use such different similarity when extending contrastive loss with generalized similarity such as t-vMF.
The proposed method was empirically shown to achieve better performance than standard regularization methods for DNNs using benchmark datasets that emulate distribution shifts.
Future work includes investigating the applicability of the proposed method for self-supervised contrastive learning.

%
%
\bibliographystyle{splncs04}
\bibliography{references}

\begin{thebibliography}{10}
\providecommand{\url}[1]{\texttt{#1}}
\providecommand{\urlprefix}{URL }
\providecommand{\doi}[1]{https://doi.org/#1}

\bibitem{bao2021predict}
Bao, Y., Chang, S., Barzilay, R.: Predict then interpolate: A simple algorithm
  to learn stable classifiers. In: International Conference on Machine
  Learning. pp. 640--650 (2021)

\bibitem{ldam}
Cao, K., Wei, C., Gaidon, A., Arechiga, N., Ma, T.: Learning imbalanced
  datasets with label-distribution-aware margin loss. Advances in neural
  information processing systems  \textbf{32} (2019)

\bibitem{carlucci2019domain}
Carlucci, F.M., D'Innocente, A., Bucci, S., Caputo, B., Tommasi, T.: Domain
  generalization by solving jigsaw puzzles. In: Computer Vision and Pattern
  Recognition conference (2019)

\bibitem{pmlr-v119-chen20j}
Chen, T., Kornblith, S., Norouzi, M., Hinton, G.: A simple framework for
  contrastive learning of visual representations. In: International Conference
  on Machine Learning. pp. 1597--1607 (2020)

\bibitem{chen2020simple}
Chen, T., Kornblith, S., Norouzi, M., Hinton, G.: A simple framework for
  contrastive learning of visual representations. In: International Conference
  on Machine Learning. pp. 1597--1607. PMLR (2020)

\bibitem{chen2021exploring}
Chen, X., He, K.: Exploring simple siamese representation learning. In:
  Computer Vision and Pattern Recognition conference. pp. 15750--15758 (2021)

\bibitem{chen2021empirical}
Chen, X., Xie, S., He, K.: An empirical study of training self-supervised
  vision transformers. In: International Conference on Computer Vision. pp.
  9640--9649 (2021)

\bibitem{randaugment}
Cubuk, E.D., Zoph, B., Shlens, J., Le, Q.V.: {RandAugment}: Practical automated
  data augmentation with a reduced search space. In: Computer Vision and
  Pattern Recognition conference. pp. 702--703 (2020)

\bibitem{arcface}
Deng, J., Guo, J., Xue, N., Zafeiriou, S.: Arcface: Additive angular margin
  loss for deep face recognition. In: Computer Vision and Pattern Recognition
  conference. pp. 4690--4699 (2019)

\bibitem{dou2019domain}
Dou, Q., Coelho~de Castro, D., Kamnitsas, K., Glocker, B.: Domain
  generalization via model-agnostic learning of semantic features. Advances in
  Neural Information Processing Systems  \textbf{32} (2019)

\bibitem{grill2020bootstrap}
Grill, J.B., Strub, F., Altch{\'e}, F., Tallec, C., Richemond, P., Buchatskaya,
  E., Doersch, C., Avila~Pires, B., Guo, Z., Gheshlaghi~Azar, M., et~al.:
  Bootstrap your own latent-a new approach to self-supervised learning.
  Advances in neural information processing systems  \textbf{33},  21271--21284
  (2020)

\bibitem{1640964}
Hadsell, R., Chopra, S., LeCun, Y.: Dimensionality reduction by learning an
  invariant mapping. In: Computer Vision and Pattern Recognition conference.
  pp. 1735--1742 (2006)

\bibitem{he2020momentum}
He, K., Fan, H., Wu, Y., Xie, S., Girshick, R.: Momentum contrast for
  unsupervised visual representation learning. In: Computer Vision and Pattern
  Recognition conference. pp. 9729--9738 (2020)

\bibitem{resnet}
He, K., Zhang, X., Ren, S., Sun, J.: Deep residual learning for image
  recognition. In: Computer Vision and Pattern Recognition conference. pp.
  770--778 (2016)

\bibitem{pmlr-v80-hu18a}
Hu, W., Niu, G., Sato, I., Sugiyama, M.: Does distributionally robust
  supervised learning give robust classifiers? In: International Conference on
  Machine Learning. pp. 2029--2037 (2018)

\bibitem{stoc_depth}
Huang, G., Sun, Y., Liu, Z., Sedra, D., Weinberger, K.Q.: Deep networks with
  stochastic depth. In: European Conference on Computer Vision. pp. 646--661
  (2016)

\bibitem{khosla2020supervised}
Khosla, P., Teterwak, P., Wang, C., Sarna, A., Tian, Y., Isola, P., Maschinot,
  A., Liu, C., Krishnan, D.: Supervised contrastive learning. Advances in
  Neural Information Processing Systems  \textbf{33},  18661--18673 (2020)

\bibitem{kim2021selfreg}
Kim, D., Yoo, Y., Park, S., Kim, J., Lee, J.: Self{R}eg: Self-supervised
  contrastive regularization for domain generalization. In: International
  Conference on Computer Vision. pp. 9619--9628 (2021)

\bibitem{adam}
Kingma, D.P., Ba, J.: Adam: A method for stochastic optimization. In:
  International Conference on Learning Representations (2015)

\bibitem{tvMF}
Kobayashi, T.: t-v{MF} similarity for regularizing intra-class feature
  distribution. In: Computer Vision and Pattern Recognition conference. pp.
  6612--6621 (2021)

\bibitem{pmlr-v139-koh21a}
Koh, P.W., Sagawa, S., Marklund, H., Xie, S.M., Zhang, M., Balsubramani, A.,
  Hu, W., Yasunaga, M., Phillips, R.L., Gao, I., Lee, T., David, E., Stavness,
  I., Guo, W., Earnshaw, B., Haque, I., Beery, S.M., Leskovec, J., Kundaje, A.,
  Pierson, E., Levine, S., Finn, C., Liang, P.: Wilds: A benchmark of
  in-the-wild distribution shifts. In: International Conference on Machine
  Learning. pp. 5637--5664 (2021)

\bibitem{KOMURA201834}
Komura, D., Ishikawa, S.: Machine learning methods for histopathological image
  analysis. Computational and Structural Biotechnology Journal  \textbf{16},
  34--42 (2018)

\bibitem{REx}
Krueger, D., Caballero, E., Jacobsen, J.H., Zhang, A., Binas, J., Zhang, D.,
  Le~Priol, R., Courville, A.: Out-of-distribution generalization via risk
  extrapolation ({RE}x). In: International Conference on Machine Learning. pp.
  5815--5826 (2021)

\bibitem{li2018deep}
Li, Y., Tian, X., Gong, M., Liu, Y., Liu, T., Zhang, K., Tao, D.: Deep domain
  generalization via conditional invariant adversarial networks. In: European
  Conference on Computer Vision. pp. 624--639 (2018)

\bibitem{liu2021just}
Liu, E.Z., Haghgoo, B., Chen, A.S., Raghunathan, A., Koh, P.W., Sagawa, S.,
  Liang, P., Finn, C.: Just train twice: Improving group robustness without
  training group information. In: International Conference on Machine Learning.
  pp. 6781--6792 (2021)

\bibitem{celaba}
Liu, Z., Luo, P., Wang, X., Tang, X.: Deep learning face attributes in the
  wild. In: International Conference on Computer Vision (December 2015)

\bibitem{loshchilovsgdr}
Loshchilov, I., Hutter, F.: Sgdr: Stochastic gradient descent with warm
  restarts. In: International Conference on Learning Representations (2017)

\bibitem{2020mmclassification}
{MMClassification Contributors}: {OpenMMLab}'s image classification toolbox and
  benchmark. \url{https://github.com/open-mmlab/mmclassification} (2020)

\bibitem{motiian2017unified}
Motiian, S., Piccirilli, M., Adjeroh, D.A., Doretto, G.: Unified deep
  supervised domain adaptation and generalization. In: International Conference
  on Computer Vision. pp. 5715--5725 (2017)

\bibitem{oord2018representation}
Oord, A.v.d., Li, Y., Vinyals, O.: Representation learning with contrastive
  predictive coding. arXiv preprint arXiv:1807.03748  (2018)

\bibitem{oren2019distributionally}
Oren, Y., Sagawa, S., Hashimoto, T.B., Liang, P.: Distributionally robust
  language modeling. In: Conference on Empirical Methods in Natural Language
  Processing and the International Joint Conference on Natural Language
  Processing. pp. 4227--4237 (2019)

\bibitem{sagawa2019distributionally}
Sagawa, S., Koh, P.W., Hashimoto, T.B., Liang, P.: Distributionally robust
  neural networks. In: International Conference on Learning Representations
  (2019)

\bibitem{7298682}
Schroff, F., Kalenichenko, D., Philbin, J.: Facenet: A unified embedding for
  face recognition and clustering. In: Computer Vision and Pattern Recognition
  Conference. pp. 815--823 (2015)

\bibitem{sohn2016improved}
Sohn, K.: Improved deep metric learning with multi-class n-pair loss objective.
  Advances in neural information processing systems  \textbf{29} (2016)

\bibitem{sun2022out}
Sun, Y., Ming, Y., Zhu, X., Li, Y.: Out-of-distribution detection with deep
  nearest neighbors. In: International Conference on Machine Learning. pp.
  20827--20840 (2022)

\bibitem{vapnik1999overview}
Vapnik, V.N.: An overview of statistical learning theory. IEEE transactions on
  neural networks  \textbf{10}(5),  988--999 (1999)

\bibitem{xie2020self}
Xie, Q., Luong, M.T., Hovy, E., Le, Q.V.: Self-training with noisy student
  improves imagenet classification. In: Computer Vision and Pattern Recognition
  conference. pp. 10687--10698 (2020)

\bibitem{pmlr-v162-zhang22z}
Zhang, M., Sohoni, N.S., Zhang, H.R., Finn, C., Re, C.: Correct-n-contrast: a
  contrastive approach for improving robustness to spurious correlations. In:
  International Conference on Machine Learning. pp. 26484--26516 (2022)

\bibitem{zhou2021examining}
Zhou, C., Ma, X., Michel, P., Neubig, G.: Examining and combating spurious
  features under distribution shift. In: International Conference on Machine
  Learning. pp. 12857--12867 (2021)

\bibitem{9847099}
Zhou, K., Liu, Z., Qiao, Y., Xiang, T., Loy, C.C.: Domain generalization: A
  survey. IEEE Transactions on Pattern Analysis and Machine Intelligence
  (2022)

\end{thebibliography}

%
%
\newpage
\appendix
\section{Additional experimental results using the CelebA dataset.} \label{appendix:celeba}
\subsection{Blond\_Hair/Gender setting}
Table~\ref{tab:res_blond_gender_overall} shows the overall accuracy of the experiments using the CelebA dataset discussed in Section~\ref{sec:res_celeba} of the main manuscript.
\begin{table}[h]
    \center
    \footnotesize
    \tabcolsep = 7pt
    \caption{Overall accuracy for the test set of the CelebA dataset (the bond\_hair/gender setting).}
    \label{tab:res_blond_gender_overall}
    \scalebox{0.9}[0.9]{
        \begin{tabular}{lllll}
            \toprule
                            & \multicolumn{2}{l}{Fixed LR} & \multicolumn{2}{l}{LR with cos decay}                                         \\
            \cmidrule(lr){2-3}\cmidrule(lr){4-5}
            {}              & Robust-DRO                   & CVaR-DRO                              & Robust-DRO        & CVaR-DRO          \\
            \midrule
            DRO-Loss        & 95.10 ($\pm$0.16)            & 94.49 ($\pm$0.47)                     & 95.28 ($\pm$0.07) & 94.36 ($\pm$0.12) \\
            \addlinespace                                                                                                                 
            +WD(1e-4)       & 93.53 ($\pm$0.79)            & 89.46 ($\pm$5.13)                     & 94.76 ($\pm$0.05) & 93.68 ($\pm$0.11) \\
            +WD(1e-3)       & 92.55 ($\pm$2.08)            & 80.35 ($\pm$11.12)                    & 93.91 ($\pm$0.17) & 92.38 ($\pm$0.18) \\
            +WD(1e-2)       & 88.80 ($\pm$3.95)            & 68.56 ($\pm$10.87)                    & 93.00 ($\pm$0.37) & 91.34 ($\pm$0.31) \\
            \addlinespace                                                                                                                 
            +SD(0.1)        & 94.93 ($\pm$0.16)            & 94.22 ($\pm$0.52)                     & 95.13 ($\pm$0.04) & 94.29 ($\pm$0.09) \\
            +SD(0.3)        & 94.51 ($\pm$0.36)            & 93.70 ($\pm$0.68)                     & 94.56 ($\pm$0.10) & 93.65 ($\pm$0.09) \\
            +SD(0.5)        & 93.80 ($\pm$0.60)            & 93.36 ($\pm$0.54)                     & 94.07 ($\pm$0.17) & 93.00 ($\pm$0.28) \\
            \addlinespace                                                                                                                 
            +RandAug        & 93.99 ($\pm$0.25)            & 92.06 ($\pm$0.75)                     & 94.20 ($\pm$0.13) & 92.11 ($\pm$0.18) \\
            \addlinespace                                                                                                                 
            +SupCon(cos)    & 94.08 ($\pm$0.54)            & 94.60 ($\pm$0.52)                     & 95.18 ($\pm$0.06) & 94.90 ($\pm$0.19) \\
            \addlinespace                                                                                                                 
            +Proposed(0.05) & 92.58 ($\pm$0.86)            & 92.57 ($\pm$0.29)                     & 94.14 ($\pm$0.18) & 93.86 ($\pm$0.25) \\
            +Proposed(0.1)  & 92.13 ($\pm$0.43)            & 91.57 ($\pm$0.78)                     & 93.23 ($\pm$0.72) & 92.41 ($\pm$0.41) \\
            +Proposed(0.15) & 90.57 ($\pm$1.46)            & 89.50 ($\pm$1.78)                     & 91.49 ($\pm$0.53) & 91.35 ($\pm$0.89) \\
            +Proposed(0.2)  & 89.55 ($\pm$2.37)            & 88.86 ($\pm$2.29)                     & 91.72 ($\pm$0.93) & 90.93 ($\pm$0.61) \\
            \bottomrule
        \end{tabular}
    }
\end{table}

\subsection{Smiling/Mouth\_Slightly\_Open setting}
In addition to the blond\_hair/gender setting explained in the main manuscript, we ran another experiment using the CelebA dataset.
We consider the problem of predicting whether a person in an image is smiling or not, and consider the attribute ``mouth\_slightly\_open'' as a confounding variable, which we refer to as the smiling/mouth\_slightly\_open setting.
Each training image belongs to one of four groups defined by smiling and mouth\_slightly\_open, whose contingency table is given in Table~\ref{tab:ct_smiling_mouth}.
\begin{table}[tbp]
    \centering
    \tabcolsep = 7pt
    \caption{Contingency tables of the CelebA dataset with respect to two attributes: smiling and mouth\_slightly\_open.}
    \label{tab:ct_smiling_mouth}
    \begin{tabular}{lrr}
        \toprule
                     & Not\_Open & Open  \\
        \midrule
        Not\_Smiling & 81339     & 23591 \\
        Smiling      & 23318     & 74351 \\
        \bottomrule
    \end{tabular}
\end{table}

\begin{figure}[tb]
    \centering
    \includegraphics[width=1.0\linewidth,clip]{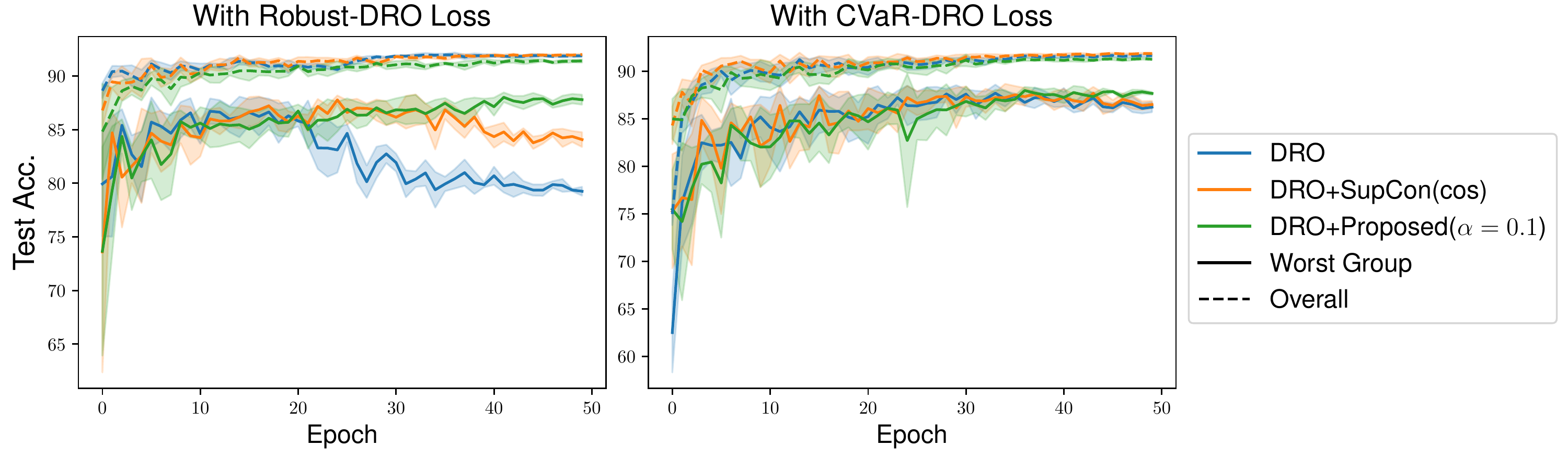}
    \caption{Test accuracy of the worst group (solid) and overall (dashed) for CelabA (the smiling/mouth\_slightly\_open setting).}
    \label{fig:celeba_smiling-mouth_r-dro_cvar-dro}
\end{figure}

Figure~\ref{fig:celeba_smiling-mouth_r-dro_cvar-dro} shows the test accuracy curves of three models: a model trained with the group DRO loss, a model trained with the group DRO loss and the SupCon loss with cosine similarity, and a model trained with the group DRO loss and the proposed loss with~$\alpha=0.1$, where the solid and dashed lines represent the worst group and overall accuracy, respectively.
We can see that the worst group accuracy deteriorates as training proceeds when the model is trained using only the DRO losses.
However, the degradation rate is moderate compared to the blond\_hair/gender setting, which is reported in the main manuscript, perhaps because there is no extreme minority in the smiling/mouth\_slightly\_open setting.

We have compared the proposed method with other DNN regularization methods. Table~\ref{tab:res_smiling_mouth_worst_group} and Table~\ref{tab:res_smiling_mouth_overall} summarizes the worst group and overall accuracy, respectively.
We can see that the proposed method achieves a better or comparable worst group accuracy compared to other regularization methods.
\begin{table}[tbph]
    \center
    \tabcolsep = 4pt
    \caption{Test accuracy of the worst group for CelebA (the smiling/mouth\_slightly\_open setting).}
    \label{tab:res_smiling_mouth_worst_group}
    \scalebox{0.9}[0.9]{
        \begin{tabular}{lllllr}
            \toprule
                             & \multicolumn{2}{l}{Fixed learning rate} & \multicolumn{2}{l}{Learning rate with decay} &                                                  \\
            \cmidrule(lr){2-3}\cmidrule(lr){4-5}
            {}               & Robust-DRO                              & CVaR-DRO                                     & Robust-DRO        & CVaR-DRO           & Average \\
            \midrule                                                                                                                                                    
            DRO-Loss         & 79.74 ($\pm$1.31)                       & 87.96 ($\pm$0.31)                            & 79.25 ($\pm$0.41) & 86.22 ($\pm$0.54)  & 83.29   \\
            \addlinespace                                                                                                                                                                        
            + WD(1e-4)       & 86.65 ($\pm$0.90)                       & 83.78 ($\pm$3.36)                            & 83.97 ($\pm$0.52) & 87.93 ($\pm$0.25)  & 85.58   \\
            + WD(1e-3)       & 83.01 ($\pm$4.56)                       & 81.20 ($\pm$6.12)                            & 88.16 ($\pm$0.29) & 87.31 ($\pm$0.25)  & 84.92   \\
            + WD(1e-2)       & 43.89 ($\pm$33.69)                      & 56.37 ($\pm$17.30)                           & 87.67 ($\pm$0.43) & 68.76 ($\pm$34.38) & 64.17   \\
            \addlinespace                                                                                                                                                                        
            + SD(0.1)        & 79.90 ($\pm$2.00)                       & 85.09 ($\pm$1.31)                            & 82.90 ($\pm$0.52) & 87.32 ($\pm$0.53)  & 83.80   \\
            + SD(0.3)        & 84.78 ($\pm$1.83)                       & 87.08 ($\pm$1.02)                            & 85.25 ($\pm$0.53) & 87.61 ($\pm$0.35)  & 86.18   \\
            + SD(0.5)        & 86.96 ($\pm$1.84)                       & 87.43 ($\pm$0.94)                            & 86.59 ($\pm$0.57) & 88.39 ($\pm$0.22)  & 87.34   \\
            \addlinespace                                                                                                                                                                        
            + RandAug        & 86.88 ($\pm$1.25)                       & 34.03 ($\pm$31.60)                           & 88.37 ($\pm$0.24) & 64.99 ($\pm$1.92)  & 68.57   \\
            \addlinespace                                                                                                                                                                        
            + SupCon(cos)    & 84.39 ($\pm$0.84)                       & 86.14 ($\pm$1.49)                            & 84.07 ($\pm$0.69) & 86.49 ($\pm$0.39)  & 85.27   \\
            \addlinespace                                                                                                                                                                        
            + Proposed(0.05) & 87.38 ($\pm$1.21)                       & 86.49 ($\pm$0.95)                            & 88.09 ($\pm$0.31) & 88.22 ($\pm$0.34)  & 87.55   \\
            + Proposed(0.1)  & 84.84 ($\pm$1.57)                       & 86.93 ($\pm$0.62)                            & 87.79 ($\pm$0.47) & 87.66 ($\pm$0.15)  & 86.81   \\
            + Proposed(0.15) & 86.03 ($\pm$0.69)                       & 84.74 ($\pm$2.11)                            & 87.07 ($\pm$0.14) & 87.42 ($\pm$0.46)  & 86.32   \\
            + Proposed(0.2)  & 85.10 ($\pm$1.48)                       & 83.01 ($\pm$3.02)                            & 86.12 ($\pm$0.63) & 86.26 ($\pm$0.69)  & 85.12   \\
            \bottomrule                                                                                                                                                    
        \end{tabular}
    }
\end{table}

\begin{table}[tbph]
    \center
    \footnotesize
    \tabcolsep = 7pt
    \caption{Overall accuracy for the test set of the CelebA dataset (the smiling/mouth\_slightly\_open setting).}
    \label{tab:res_smiling_mouth_overall}
    \scalebox{0.9}[0.9]{
        \begin{tabular}{lllll}
            \toprule
                            & \multicolumn{2}{l}{Fixed LR} & \multicolumn{2}{l}{LR with cos decay}                                          \\
            \cmidrule(lr){2-3}\cmidrule(lr){4-5}
            {}              & Robust-DRO                   & CVaR-DRO                              & Robust-DRO        & CVaR-DRO           \\
            \midrule                                                                                                                         
            DRO-Loss        & 91.74 ($\pm$0.13)            & 91.12 ($\pm$0.21)                     & 91.88 ($\pm$0.07) & 91.63 ($\pm$0.13)  \\
            \addlinespace                                                                                                                                                
            +WD(1e-4)       & 90.80 ($\pm$0.23)            & 89.78 ($\pm$1.18)                     & 90.72 ($\pm$0.07) & 91.03 ($\pm$0.08)  \\
            +WD(1e-3)       & 89.90 ($\pm$1.91)            & 88.45 ($\pm$2.52)                     & 91.34 ($\pm$0.14) & 90.89 ($\pm$0.06)  \\
            +WD(1e-2)       & 73.62 ($\pm$14.37)           & 80.77 ($\pm$5.82)                     & 90.88 ($\pm$0.21) & 81.93 ($\pm$15.98) \\
            \addlinespace                                                                                                                                             
            +SD(0.1)        & 91.70 ($\pm$0.18)            & 90.82 ($\pm$0.56)                     & 92.21 ($\pm$0.07) & 91.56 ($\pm$0.12)  \\
            +SD(0.3)        & 90.96 ($\pm$0.23)            & 91.18 ($\pm$0.40)                     & 91.28 ($\pm$0.11) & 91.71 ($\pm$0.07)  \\
            +SD(0.5)        & 91.72 ($\pm$0.23)            & 91.25 ($\pm$0.25)                     & 92.12 ($\pm$0.05) & 91.65 ($\pm$0.05)  \\
            \addlinespace                                                                                                                                           
            +RandAug        & 91.81 ($\pm$0.27)            & 60.62 ($\pm$15.32)                    & 91.81 ($\pm$0.07) & 68.73 ($\pm$4.82)  \\
            \addlinespace                                                                                                                                           
            +SupCon(cos)    & 91.71 ($\pm$0.13)            & 91.23 ($\pm$0.45)                     & 92.00 ($\pm$0.08) & 91.85 ($\pm$0.11)  \\
            \addlinespace                                                                                                                                             
            +Proposed(0.05) & 91.15 ($\pm$0.30)            & 91.26 ($\pm$0.37)                     & 91.71 ($\pm$0.06) & 91.61 ($\pm$0.08)  \\
            +Proposed(0.1)  & 90.91 ($\pm$0.70)            & 90.84 ($\pm$0.23)                     & 91.40 ($\pm$0.15) & 91.27 ($\pm$0.09)  \\
            +Proposed(0.15) & 90.21 ($\pm$0.38)            & 90.21 ($\pm$0.79)                     & 90.88 ($\pm$0.17) & 91.07 ($\pm$0.31)  \\
            +Proposed(0.2)  & 89.80 ($\pm$0.63)            & 88.97 ($\pm$1.40)                     & 90.39 ($\pm$0.57) & 90.34 ($\pm$0.80)  \\
            \bottomrule
        \end{tabular}
    }
\end{table}

\section{Experiments with the iWildCam dataset} \label{appendix:iwildcam}
In addition to the experiments on the Camelyon17 dataset, we conducted experiments on the iWildCam2020-WILDS~(v2.0) dataset~\cite{pmlr-v139-koh21a}, which we will refer to as iWildCam for simplicity, to evaluate the proposed method in the domain generalization situation.
The iWildCam dataset consists of photos taken by camera traps placed in the wild to monitor animal populations.
Each photo is labeled with one of 182 classes that indicate the animal species, including the label that indicates no animal in the photo.
The location of each camera trap is considered a domain, and the iWildCam dataset contains photos from 323 different domains.
The iWildCam dataset has five splits\footnote{These splits are named train, id\_val, id\_test, val, and test in the original data.} including~$\bar{\mathcal{D}}_\text{train}$, $\bar{\mathcal{D}}_\text{val}^\text{ID}$, $\bar{\mathcal{D}}_\text{test}^\text{ID}$, $\bar{\mathcal{D}}_\text{test}^\text{OOD1}$, and~$\bar{\mathcal{D}}_\text{test}^\text{OOD2}$.
The domains of~$\bar{\mathcal{D}}_\text{train}$, $\bar{\mathcal{D}}_\text{val}^\text{ID}$, and~$\bar{\mathcal{D}}_\text{test}^\text{ID}$ are the same, and $\bar{\mathcal{D}}_\text{test}^\text{OOD1}, \bar{\mathcal{D}}_\text{test}^\text{OOD2}$ are data from different domains.

\begin{figure}[htbp]
    \centering
    \includegraphics[width=1.0\linewidth,clip]{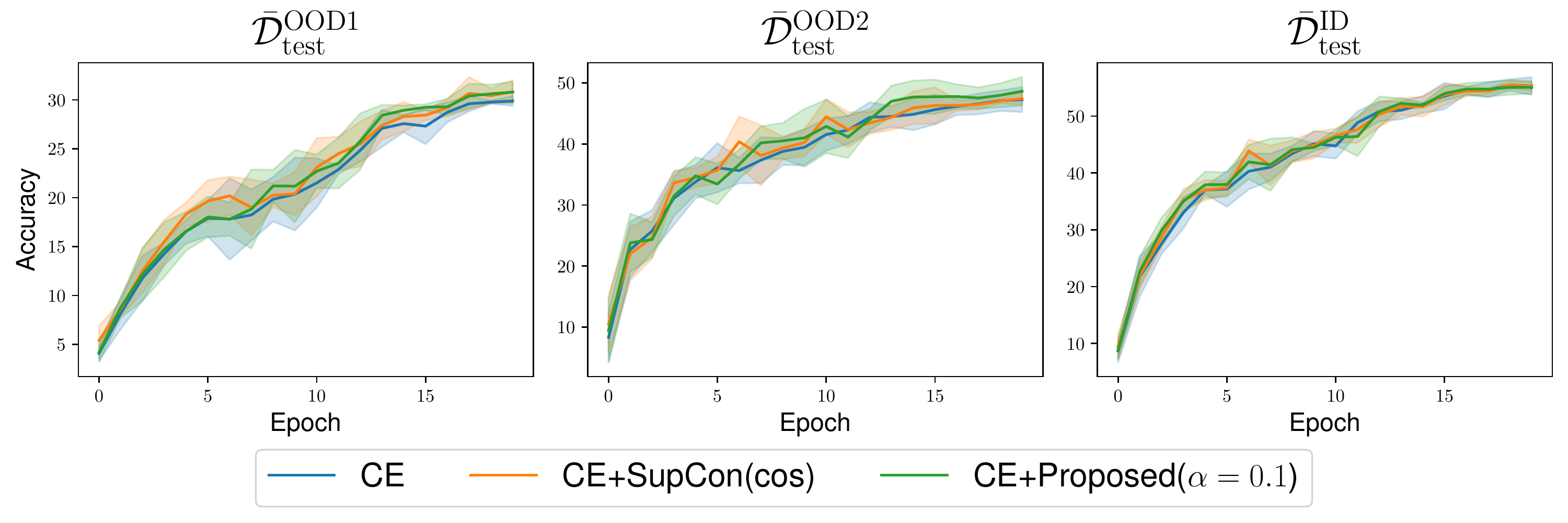}
    \caption{Accuracy on the test splits of the iWildCam dataset.}
    \label{fig:iwildcam_vs_cos}
\end{figure}

We trained a model using~$\bar{\mathcal{D}}_\text{train}$, where the model and training settings are the same as for the Camelyon17 dataset except for the training epochs of~20.
Figure~\ref{fig:iwildcam_vs_cos} shows the test accuracy curves during training on each of $\bar{\mathcal{D}}_\text{test}^\text{OOD1}$, $\bar{\mathcal{D}}_\text{test}^\text{OOD2}$, and~$\bar{\mathcal{D}}_\text{test}^\text{ID}$.
Compared to the results of the Camelyon17 dataset, the performance on OOD splits does not degrade on the iWildCam dataset.
The reason for this is not clear, but the fact that the training split of iWildCam includes samples from many domains compared to Camelyon17 may have led to the prevention of overfitting, as discussed in the main manuscript.
Table~\ref{tab:res_iwildcam} shows the accuracy on each test split at the end of training.
We can see that the proposed method achieves slightly better performance on OOD splits compared to training with CE.

\begin{table}[htbp]
    \center
    \footnotesize
    \tabcolsep = 7pt
    \caption{Accuracy for the test sets of the iWildCam dataset.}
    \label{tab:res_iwildcam}
    \begin{tabular}{lllll}
        \toprule
        {}                        & $\bar{\mathcal{D}}_\text{test}^\text{OOD1}$ & $\bar{\mathcal{D}}_\text{test}^\text{OOD2}$ & $\bar{\mathcal{D}}_\text{test}^\text{ID}$ \\
        \midrule                                                                   
        CE                        & 29.89 ($\pm$0.51)                           & 47.25 ($\pm$2.05)                           & 55.31 ($\pm$1.58)                         \\
        CE+SupCon(cos)            & 30.83 ($\pm$1.22)                           & 47.44 ($\pm$1.04)                           & 55.25 ($\pm$0.74)                         \\
        CE+Proposed($\alpha=0.1$) & 30.79 ($\pm$1.07)                           & 48.63 ($\pm$2.33)                           & 55.05 ($\pm$1.10)                         \\
        \bottomrule
    \end{tabular}
\end{table}

\section{Additional experimental results using the Camelyon17 dataset.} \label{appendix:camelyon17}
\begin{figure}[tbp]
    \centering
    \begin{minipage}[b]{\hsize}
        \centering
        \includegraphics[width=1.0\linewidth,clip]{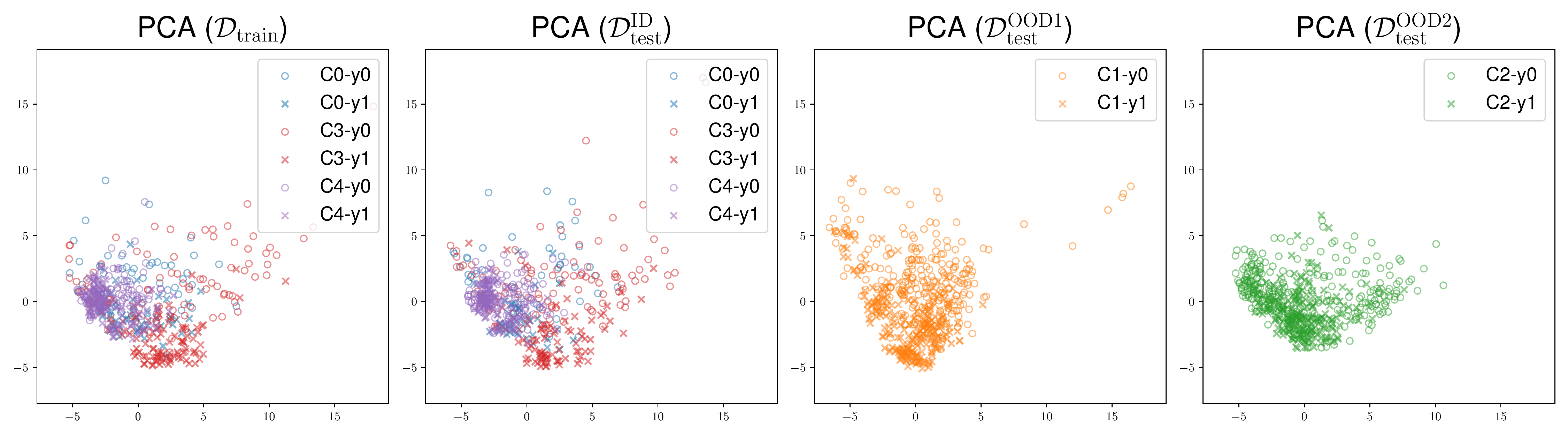}
        \subcaption{Initial model.}
    \end{minipage}
    \begin{minipage}[b]{\hsize}
        \centering
        \includegraphics[width=1.0\linewidth,clip]{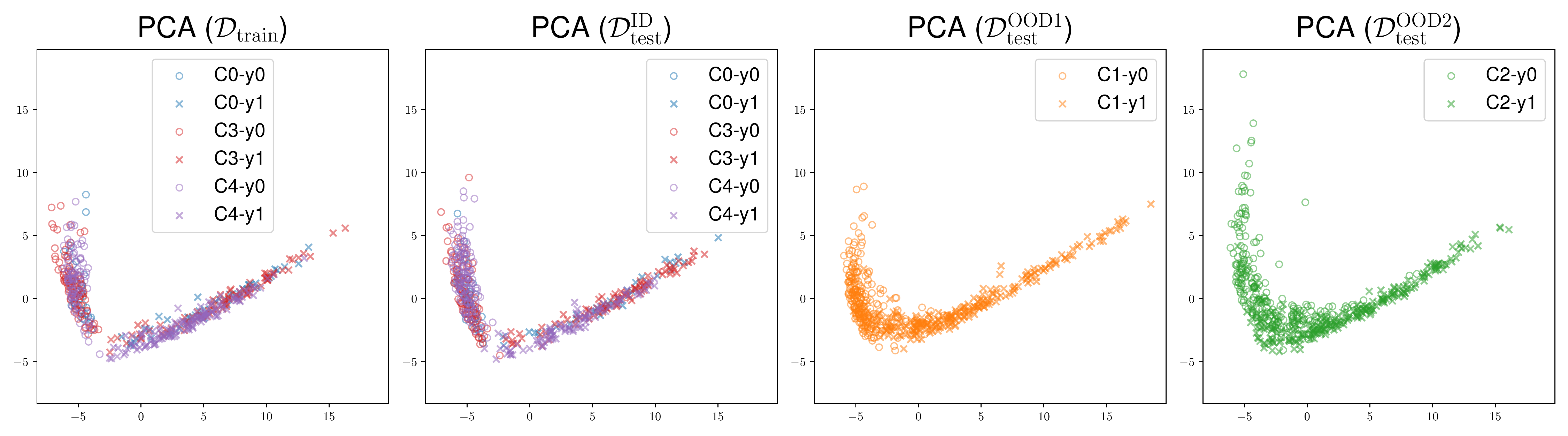}
        \subcaption{Model trained with CE.}
    \end{minipage}
    \begin{minipage}[b]{\hsize}
        \centering
        \includegraphics[width=1.0\linewidth,clip]{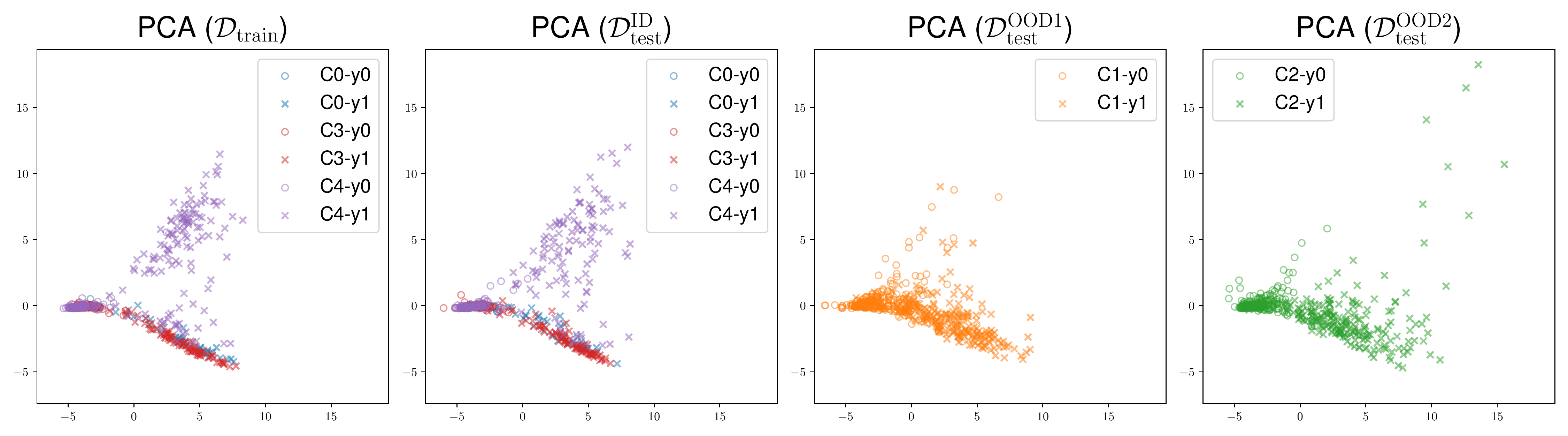}
        \subcaption{Model trained with CE+SupCon(cos).}
    \end{minipage}
    \begin{minipage}[b]{\hsize}
        \centering
        \includegraphics[width=1.0\linewidth,clip]{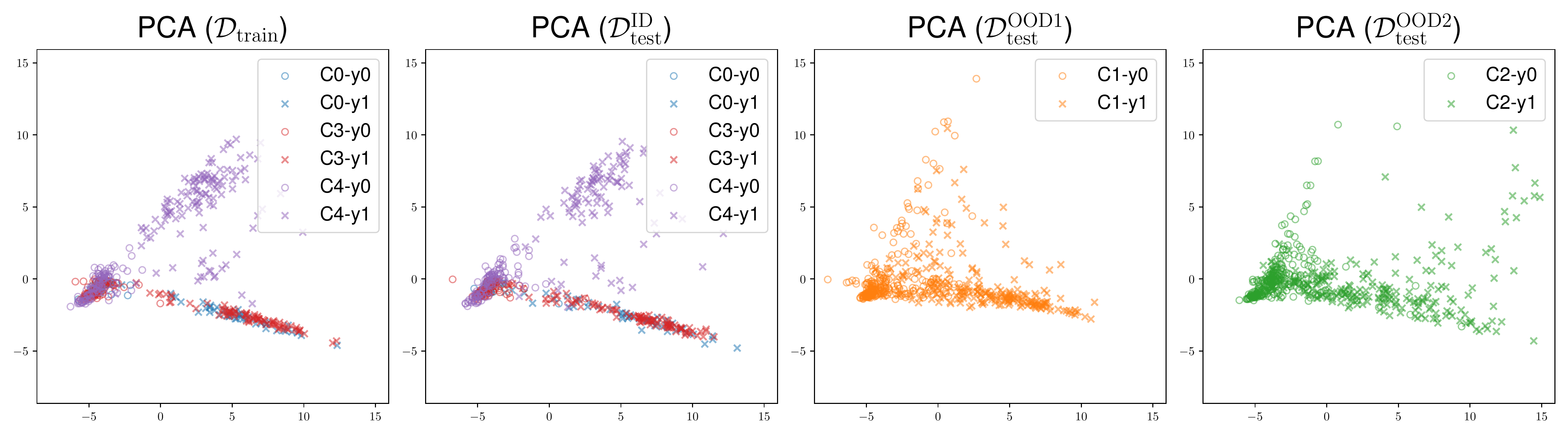}
        \subcaption{Model trained with CE+Proposed($\alpha=0.1$).}
    \end{minipage}
    \caption{PCA visualization of embeddings for the Camelyon17 dataset. ``C'' and ``y'' in the legend indicate center and label, respectively.}
    \label{fig:pca_camelyon17}
\end{figure}

Figure~\ref{fig:pca_camelyon17} shows the PCA visualization of the embeddings for samples in~$\mathcal{D}_{\text{train}}$, $\mathcal{D}_{\text{test}}^{\text{ID}}$, $\mathcal{D}_{\text{test}}^{\text{OOD1}}$, and~$\mathcal{D}_{\text{test}}^{\text{OOD2}}$ of the Camelyon17 dataset, where embeddings are obtained using each of four models: The initial model, a model trained with CE, a model trained with CE and SupCon with cosine similarity, and a model trained with CE and the proposed loss with~$\alpha=0.1$.
The point colors correspond to the hospital where the sample was obtained, and the ``o'' and ``x'' marks mean~$y=0$ (no tumor) and~$y=1$ (tumor).

\end{document}